\documentclass{article}

\usepackage{amsmath,graphicx}
\usepackage{amsmath,graphicx,caption}
\usepackage{amssymb,amsmath,latexsym}
\usepackage{multirow}
\usepackage{footnote,xspace,syntonly}
\usepackage{cite,color,epsfig,verbatim}
\usepackage{epstopdf}
\usepackage{algorithm}
\usepackage{algpseudocode}
\usepackage{float}
\usepackage{color}
\usepackage{varwidth}
\usepackage[keeplastbox]{flushend}
\usepackage{cite}
\usepackage{fixltx2e}

\usepackage{graphicx}

\usepackage{subcaption}              

\usepackage[utf8]{inputenc}

\usepackage{amsfonts}
\usepackage{amsmath}
\usepackage{mathtools}

\usepackage{amssymb}

\usepackage{bm}

\usepackage{color,verbatim}
\usepackage{multirow}
\usepackage{accents}

\usepackage{theoremref}

\usepackage{flushend}

\usepackage{url}

\newcounter{rulecounter}
\newcommand{\resetrule}{ \setcounter{rulecounter}{0}}
\resetrule

\newsavebox{\selvestebox}
\newenvironment{colbox}[1]
  {\newcommand\colboxcolor{#1}%
   \begin{lrbox}{\selvestebox}%
   \begin{minipage}{\dimexpr\columnwidth-2\fboxsep\relax}}
  {\end{minipage}\end{lrbox}%
   \begin{center}
   \colorbox{\colboxcolor}{\usebox{\selvestebox}}
   \end{center}}

\definecolor{orange}{rgb}{1,0.8,0}
\definecolor{gray}{rgb}{.9,0.9,0.9}
\definecolor{darkgray}{rgb}{.3,0.3,0.3}
\definecolor{darkblue}{rgb}{.1,0.0,0.3}
\definecolor{lightblue}{rgb}{0.7,0.7,1}
\definecolor{lightred}{rgb}{1,0.7,.7}
\definecolor{purple}{RGB}{204,153,255}
\definecolor{lightgray}{rgb}{.95,0.95,0.95}
\definecolor{lightgreen}{rgb}{0.3,0.5,0.3}
\definecolor{darkgreen}{rgb}{0.05,0.3,0.05}

\newtheorem{myproposition}{Proposition}
\newtheorem{myremark}{Remark}
\newtheorem{myproblemstatement}{Problem Statement}
\newtheorem{mylemma}{Lemma}
\newtheorem{mytheorem}{Theorem}

\newtheorem{mycorollary}{Corollary}

\usepackage{pgfplots}
\pgfplotsset{compat=newest}
\usetikzlibrary{plotmarks}
\usetikzlibrary{arrows.meta}
\usepgfplotslibrary{patchplots}
\usepackage{grffile}

\pgfplotsset{plot coordinates/math parser=false}
\newlength\mywidth
\newlength\myheight
\definecolor{mycolorBL1}{rgb}{0.00000,0.54700,0.54100}%
\definecolor{mycolorBL2}{rgb}{0.85000,0.92500,0.09800}%
\definecolor{mycolorBL3}{rgb}{0.92900,0.69400,0.72500}%
\definecolor{mycolor1}{rgb}{0.00000,0.44700,0.74100}%
\definecolor{mycolor2}{rgb}{0.85000,0.32500,0.09800}%
\definecolor{mycolor3}{rgb}{0.92900,0.69400,0.12500}%
\definecolor{mycolor4}{rgb}{0.89400,0.18400,0.15600}%
\definecolor{mycolor5}{rgb}{0.46600,0.67400,0.18800}%
\definecolor{mycolor6}{rgb}{0.30100,0.74500,0.93300}%
\definecolor{mycolor7}{rgb}{0.63500,0.07800,0.18400}%
\definecolor{colorJSIGoT}{rgb}{1,0.0032500,0.001}%
\newcommand{\xlabelfontsize}{\normalsize}
\newcommand{\ylabelfontsize}{\normalsize}
\newcommand{\legendfontsize}{\normalsize}
\newcommand{\ticklabelfontsize}{\scriptsize}

\newtheorem{lemma}{Lemma}

\newtheorem{theorem}{Theorem}

\newtheorem{remark}{Remark}

\def\bbtheta{{\mbox{\boldmath $\theta$}}}

\def\bbSig{{\mbox{\boldmath $\Sigma$}}}
\def\bbphi{{\mbox{\boldmath $\phi$}}}
\def\bbPhi{{\mbox{\boldmath $\Phi$}}}

\def\bbzeta{{\mbox{\boldmath $\zeta$}}}

\newif\ifshowtikz
\showtikztrue
\allowdisplaybreaks[4]

\let\oldtikzpicture\tikzpicture
\let\oldendtikzpicture\endtikzpicture

\renewenvironment{tikzpicture}{%
\ifshowtikz\expandafter\oldtikzpicture%
\else\comment%
\fi
}{%
\ifshowtikz\oldendtikzpicture%
\else\endcomment%
\fi}

\title{Robust and Adaptive Temporal-Difference Learning  Using  An Ensemble of Gaussian Processes}

\author{ 
	Qin Lu\textsuperscript{\rm 1} ~~
Georgios B. Giannakis\textsuperscript{\rm 1} \\	 
	\textsuperscript{\rm 1}\textit{University of Minnesota} \\
	\texttt{\{qlu, georgios\}@umn.edu} \\
}

\begin{document}
	
	\maketitle
\begin{abstract}
Value function approximation is a crucial module for policy evaluation in reinforcement learning when the state space is large or continuous. The present paper takes a generative perspective on policy evaluation via temporal-difference (TD) learning, where a Gaussian process (GP) prior is presumed on the sought value function, and instantaneous rewards are probabilistically generated based on value function evaluations at two consecutive states. Capitalizing on a random feature-based approximant of the GP prior, an online scalable (OS) approach, termed {OS-GPTD}, is developed to estimate the value function for a given policy by observing a sequence of state-reward pairs. To benchmark the performance of OS-GPTD even in an adversarial setting, where the modeling assumptions are violated, complementary worst-case analyses are performed by upper-bounding the cumulative Bellman error as well as the long-term reward prediction error, relative to their counterparts from a fixed value function estimator with the entire state-reward trajectory in hindsight. Moreover, to alleviate the limited expressiveness associated with a single fixed kernel, a 
weighted ensemble (E) of GP priors is employed to yield an alternative scheme, termed OS-EGPTD, that can jointly infer the value function, and select interactively the EGP kernel on-the-fly. Finally, performances of the novel OS-(E)GPTD schemes are evaluated on two benchmark problems.
\end{abstract}

\section{Introduction} \label{sec:intro}

In reinforcement learning (RL), an agent learns a policy function yielding as output a sequence of actions with input the observed states of an environment, so that a cumulative reward criterion is maximized. As a dynamic system, the environment can be \emph{unknown, nonlinear, and complex}, while the agent has to learn how to act directly from state-reward observations. RL has documented success in several disciplines, including e.g., control, information theory, operations research, machine learning, and statistics, to list a few \cite{sutton2018reinforcement}. 

Many algorithms for obtaining an optimal policy achieving the highest cumulative reward from each state are based on the  policy iteration method \cite{book2014puterman}, in which the so-termed value function must be estimated for a sequence of fixed policies, making \emph{value function estimation} (policy evaluation) a critical algorithmic component. Temporal-difference (TD) learning \cite{sutton2018reinforcement}, is arguably the best known method for value function estimation. It relies on forming the so-called TDs to update an estimate of the value function on-the-fly.
Although intuitively simple, the classic TD learning algorithm can be intractable when the state space of the environment is large or continuous, often the case in contemporary control and artificial intelligence tasks \cite{2015drl}. A scalable practice to overcome this hurdle leverages approximants of the value function, including e.g., linear ones \cite{sutton2018reinforcement}, deep neural networks \cite{2015drl}, or Gaussian processes (GPs) \cite{engel2003bayes,kuss2004gaussian}. 

Speaking of GPs, cross-fertilizing the merits of kernel
learning and Bayesian methods, they have been widely employed in several learning tasks, including regression, classification, clustering, and dimensionality reduction, in addition to RL
\cite{williams2006gaussian}. Postulating a probabilistic generative model, GP-based methods not only allow learning over a rich class of nonlinear functions, but also quantify the associated uncertainty. Leveraging GPs for RL has led to efforts in model-based approaches including \cite{kuss2004gaussian, 
deisenroth2011pilco,
pami2017,dgpirl}, as well as model-free ones, as in \cite{engel2003bayes,engel2005reinforcement,
grande2014sample,gp4metarl}. Specifically in \cite{engel2003bayes}, the developed GPTD algorithm estimates the value function from a Bayesian perspective, where the value function is assumed drawn from a GP prior, and a probabilistic observation model is postulated to generate the instantaneous reward from value function realizations at two consecutive states.
Further, a revised observation model has been proposed in \cite{engel2005reinforcement}, along with a GP-based SARSA algorithm.

Concerning \emph{performance analysis} of TD methods, several attempts have been made.
Under some probabilistic assumptions about the environment to effect a sequence of Markovian observations, or of samples from a static Markov decision process (MDP), both asymptotic and non-asymptotic analyses of TD learning with function approximation have been widely studied; see e.g., \cite{tac1997td,maei2009td, dalal2018td,bhandari2018td,srikant2019td
}. Without any statistical assumptions about the environment,
worst-case analysis on the long-term reward predictive error has been reported for TD learning and its variants in an online setting
\cite{schapire1996worst, li2008worst}.  Further along this line, connections between online long-term reward prediction error and
Bellman error were drawn in \cite{sun2015online}.
However, their results do not directly apply to GPTD-type algorithms. Among other GP-based RL frameworks, sample complexity analysis has been conducted for both model-based and model-free methods in \cite{grande2014sample}. 

{\bf Contributions.} Building on the GPTD framework for value estimation, the present work develops \emph{online scalable} (OS) schemes whose performance can be also quantified using worst-case analyses. Relative to past works, the contributions can be summarized as follows.\footnote{Part of this work is published in\cite{lu2021gaussian}.}

\begin{enumerate}
	\item[{\bf c1)}] 
Leveraging the random feature (RF)  approximant of the GP prior,  we develop an OS-GPTD scheme, which 
readily accommodates streaming reward-state observations to learn the value function on-the-fly.

	\item[{\bf c2)}] To benchmark the performance of OS-GPTD even in the adversarial setting where state-reward pairs can be chosen adversarially to violate the generative modeling assumptions,
	two complementary worst-case analyses are provided.
	The first analysis establishes the \emph{no-regret} property of  the online function estimator given by OS-GPTD in terms of Bellman error, while the second one further bounds the accumulated future reward prediction error incurred by OS-GPTD compared with that corresponding to the best fixed function estimator with data in hindsight.

	\item[{\bf c3)}] To enrich the expressiveness of the sought value function, 
	a weighted ensemble (E-) of OS-GPTDs from multiple GP experts with different priors, 
	are employed to not only infer the value function, but also select the appropriate kernel(s) adaptively. 
\end{enumerate}

\section{Preliminaries}

To elicit the main contributions of this paper, some preliminaries on RL and GPTD learning in particular, are outlined in this section.

Characterizing the interactions between an agent and the environment, an RL problem is generally formulated as a discounted MDP specified by the quintuple $\mathcal{M}:=\{\mathcal{S}, \mathcal{A}, \mathcal{P}, {R}, \gamma\}$, where $\mathcal{S}$ is the state space of the environment, $\mathcal{A}$ is the action space of the agent, $\mathcal{P}$ defines the state dynamics, ${R}$ is the reward function, and $\gamma$ is the discount factor belonging to $[0,1)$. 

Capitalizing on simulated or real-world experiences, the objective of the agent is to obtain an optimal policy that maximizes the cumulative rewards 
from each state. To this end, policy- and value-based approaches have been devised; see, e.g., \cite{book2014puterman}. The state value function is defined as
\begin{equation}
v(\mathbf{s}) := \mathbb{E}\!\left[\sum_{\tau =1}^\infty \gamma^{\tau-1} r_{\tau} | \mathbf{s}_1 = \mathbf{s}\right]
\end{equation}
where the expectation is taken over the entire MDP trajectory conditioned on the initial state $\mathbf{s}_1=\mathbf{s}\in\mathcal{S}$, while following some fixed policy to take future actions.
Value-based methods proceed by alternating between value function estimation for a given policy, and policy improvement using the predicted value function. Value function estimation relies on the popular TD-based iteration indexed by superscript $i$ with scalar $\alpha>0$~\cite{sutton2018reinforcement}
\begin{equation}
v^{(i+1)}(\mathbf{s}_{t}) = v^{(i)}(\mathbf{s}_{t}) + \alpha\!\left[r_{t}+\gamma v^{(i)}(\mathbf{s}_{t+1}) - v^{(i)}(\mathbf{s}_{t})\right] \label{eq:TD_tabular}.
\end{equation}
When the state space becomes large or continuous, the computational requirements of value function estimation in \eqref{eq:TD_tabular} are overwhelming as all solvers of Bellman's equation. This has motivated scalable approaches leveraging low-dimensional approximants of $v$, including linear \cite{sutton2018reinforcement}, deep neural networks \cite{2015drl}, and Bayesian ones~\cite{engel2003bayes}. 

In Bayesian approximators, $v$ is assumed drawn from a GP prior denoted as $v\sim \mathcal{GP}(0,\kappa(\mathbf{s},\mathbf{s}'))$, where $\kappa$ is a kernel function measuring the pairwise similarity between the function values at $d$-dimensional states $\mathbf{s}$ and $\mathbf{s}'$. The GP prior on $v$ implies that the joint probability density function (pdf) for any number of function values $\mathbf{v}_t:=[v(\mathbf{s}_1)~v(\mathbf{s}_2)~ \cdots~ v(\mathbf{s}_t)]^\top$
($^\top$ stands for transposition) evaluated
at $\mathbf{S}_t:= [\mathbf{s}_1~ \mathbf{s}_2~ \cdots~  \mathbf{s}_t]$ is multivariate Gaussian \cite{williams2006gaussian}
\begin{equation}
p(\mathbf{v}_t| \mathbf{S}_t) = \mathcal{N}(\mathbf{v}_t; \mathbf{0}_t,\mathbf{K}_t) \label{eq:GP_Prior}
\end{equation}
where $\mathbf{0}_t$ is the $t$-dimensional all-zero vector, and
the $(i,j)$-th entry of 
$\mathbf{K}_t\in\mathbb{R}^{t\times t}$ denotes the covariance between $v(\mathbf{s}_{i})$ and $v(\mathbf{s}_{j})$; that is, $[\mathbf{K}_t]_{i,j}:={\rm cov}(v(\mathbf{s}_{i}),v(\mathbf{s}_{j})) = \kappa(\mathbf{s}_{i},\mathbf{s}_{j})$.

Upon convergence, the update in \eqref{eq:TD_tabular} implies that the term in square brackets must vanish. This in turn suggests a generative model relating the reward $r_t$ with the value function $v(\mathbf{s})$ as~\cite{engel2003bayes}
\begin{equation}
r_{t} = v(\mathbf{s}_{t})-\gamma v(\mathbf{s}_{t+1}) + n_{t} \label{eq: r_t}
\end{equation}
where the noise $n_{t}$ is assumed white, uncorrelated with any $v(\mathbf{s})$, and zero-mean Gaussian with variance $\sigma_n^2$.

Starting from some initial state $\mathbf{s}_1\in\mathcal{S}$, taking subsequent actions by following a given policy generates a trajectory of states and rewards,
namely $\{ \mathbf{s}_1, r_1, \mathbf{s}_2, r_2,$ $ \mathbf{s}_3, \ldots\}$. Upon collecting the rewards 
into vector $\mathbf{r}_t := [r_1~ r_2~ \cdots~ r_t]^\top$, 
the conditional likelihood is
\begin{equation}
p(\mathbf{r}_t | \mathbf{v}_{t+1}, \mathbf{S}_{t+1}) = \mathcal{N}(\mathbf{r}_t; \mathbf{H}_t \mathbf{v}_{t+1},  \sigma^2_n \mathbf{I}_t)  \label{eq:likelihood}
\end{equation}
where $\mathbf{I}_t$ is the $t\times t$ identity matrix and the $t\times (t+1)$ observation matrix is 
\begin{equation}
\mathbf{H}_t := 
\begin{bmatrix}
1 & -\gamma & 0 & \cdots & 0\\
0 & 1 & -\gamma & \cdots & 0 \\
\vdots &  & \ddots & \ddots & \vdots \\
0 & 0 & \cdots & 1 & -\gamma
\end{bmatrix}. \nonumber
\end{equation}
Given the GP prior in \eqref{eq:GP_Prior} and the conditional likelihood in \eqref{eq:likelihood} for $\mathbf{r}_t = \mathbf{H}_t \mathbf{v}_{t+1} + \mathbf{n}_t$,  the goal is to predict the pdf of $v(\cdot)$ at some unseen state $\mathbf{s}$. With this goal in mind, let us start off by writing the joint pdf of $\mathbf{r}_t$ and $v(\mathbf{s})$ as 
\begin{align}
&\begin{bmatrix}     \mathbf{r}_t \\  v(\mathbf{s}) \end{bmatrix}\sim  \mathcal{N}\!\left(\!
\mathbf{0}_{t+1},
\begin{bmatrix} 
\mathbf{Q}_t & \mathbf{H}_t \mathbf{k}_{t+1} (\mathbf{s}) \\
\mathbf{k}_{t+1}^\top (\mathbf{s})\mathbf{H}_t^\top & \kappa(\mathbf{s},\mathbf{s})
\end{bmatrix}
\!   \right)   \label{eq:joint_pdf}
\end{align}
where $\mathbf{Q}_t = \mathbf{H}_t\mathbf{K}_{t+1}\mathbf{H}_t^\top + \sigma^2_n \mathbf{I}_t $ is the covariance matrix of the marginal pdf 
\begin{align}
p(\mathbf{r}_t|\mathbf{S}_{t+1}) &= \int p(\mathbf{r}_t|\mathbf{v}_{t+1})p(\mathbf{v}_{t+1}|\mathbf{S}_{t+1})d\mathbf{v}_{t+1} \nonumber\\
&=\mathcal{N}(\mathbf{r}_t; \mathbf{0}_t, \mathbf{Q}_t)\nonumber
\end{align}
and $\mathbf{H}_t \mathbf{k}_{t+1} (\mathbf{s})$ denotes the cross-covariance 
\begin{equation}
\mathbb{E}[\mathbf{r}_t v(\mathbf{s})|\mathbf{S}_{t+1}] = \mathbb{E}[\mathbf{H}_t\mathbf{v}_{t+1} v(\mathbf{s})|\mathbf{S}_{t+1}] = \mathbf{H}_t \mathbf{k}_{t+1} (\mathbf{s}) \nonumber
\end{equation}
with $[\mathbf{k}_{t+1} (\mathbf{s})]_{\tau} := \kappa(\mathbf{s}_{\tau},\mathbf{s})$ for $\tau=1,2,\ldots, t+1$.

Leveraging the pdf \eqref{eq:joint_pdf}, the conditional pdf of $v(\mathbf{s})$ is 
\begin{equation}
p(v(\mathbf{s})| \mathbf{r}_t, \mathbf{S}_{t+1}) = \mathcal{N} (v(\mathbf{s}); \hat{v}(\mathbf{s}), \sigma_v^2 (\mathbf{s}))   \label{eq:batch_estimator}
\end{equation}
with mean and variance given by
\begin{subequations}
	\begin{align}
		\hspace*{-0.1cm}
	\hat{v}(\mathbf{s}) &=  \mathbf{k}_{t+1}^\top (\mathbf{s})\mathbf{H}_t^\top \mathbf{Q}_t^{-1}\mathbf{r}_t  \label{eq:mean_GP}\\
	\sigma_v^2 (\mathbf{s}) &=  \kappa(\mathbf{s}, \mathbf{s})\! - \!  \mathbf{k}_{t+1}^\top (\mathbf{s}) \mathbf{H}_t^\top \mathbf{Q}_t^{-1} \mathbf{H}_t \mathbf{k}_{t+1} (\mathbf{s}). \label{eq:var_GP}
	\end{align}	
\end{subequations}

At negligible extra cost, the accompanying variance \eqref{eq:var_GP} not only provides a self-assessment of the accuracy of estimate \eqref{eq:mean_GP}, but can also guide exploration for policy improvement.
Clearly, evaluating the function \eqref{eq:batch_estimator} incurs computational complexity $\mathcal{O}(t^3)$ and storage complexity  $\mathcal{O}(t^2)$, which can become prohibitive as 
$t$ grows. Although a scalable solution has been devised for the online GPTD algorithm in \cite{engel2003bayes}, the estimation performance is compromised by the information loss incurred by the sparsification employed. 

Our pursuit in the rest of this paper aims at online scalable solvers and their associated performance analyses.

\section{Scalable TD learning with a single GP}
\label{sec:Single_GP}
Several attempts have been made to effect scalability in GP-based learning; see, e.g., \cite{ quinonero2005unifying, quia2010sparse
}. For GPTD in particular,
a sparsification scheme was advocated in \cite{engel2003bayes}; and a scalable value function estimator was reported in \cite{martin2018sparse} using an approximant \cite{quinonero2005unifying}, which summarizes the training data through a reduced number of pseudo data with inducing inputs. 
Unfortunately, performance of both low-rank approximants of  $\mathbf{K}_t$ has 
not been analyzed.

Aiming at a low-rank approximant with quantifiable performance, we rely here on a {\it shift-invariant, real, and even} $\bar{\kappa} (\cdot)$, which is given by the inverse Fourier transform of a power spectral density $\pi_{\bar{\kappa}}$; that is,
\begin{equation}
\bar{\kappa}(\mathbf{s}, \mathbf{s}') = \bar{\kappa}(\mathbf{s}-\mathbf{s}') = \int \pi_{\bar{\kappa}} (\bbzeta) e^{j\bbzeta^\top (\mathbf{s} - \mathbf{s}')} d\bbzeta . \label{eq:kappa_SI}
\end{equation}
If $\bar{\kappa} (\cdot)$ is standardized so that $\bar{\kappa}(\mathbf{0}) = \int \pi_{\bar{\kappa}} (\bbzeta) d\bbzeta = 1$, then $\pi_{\bar{\kappa}}$ boils down to a pdf. Thus, the integral in \eqref{eq:kappa_SI} can be expressed as the expectation
\begin{equation}
\bar{\kappa}(\mathbf{s}, \mathbf{s}')  \!= \! \mathbb{E}_{\pi_{\bar{\kappa}}}\! \left[e^{j\bbzeta^\top (\mathbf{s} - \mathbf{s}')} \right]  =  \mathbb{E}_{\pi_{\bar{\kappa}}}\!\left[ \cos\!\left( \bbzeta^\top (\mathbf{s} - \mathbf{s}') \right) \right]. \nonumber 
\end{equation}
Upon drawing a sufficiently large number $D$ ($\gg d$), of independent and identically distributed (i.i.d.) samples (a.k.a. features) $\{\bbzeta_i \}_{i = 1}^D$ from $\pi_{\bar{\kappa}}$, the last expectation over $\bbzeta$ can be approximated by the sample average, \footnote{Quantities with $\check{(\cdot)}$ involve the RF approximation.}
which can be further rewritten as 
\begin{equation}
\check{\bar{\kappa}}(\mathbf{s}, \mathbf{s}') = \bbphi_{\zeta}^{\top} (\mathbf{s})\bbphi_{\zeta}(\mathbf{s}') \label{kern_est}
\end{equation}
where $\bbphi_{\zeta} (\mathbf{s})$ is the real $2D\times 1$ random feature (RF) vector given by~\cite{rahimi2008random}
\begin{align}	\vspace{-.4cm}
&\bbphi_{\zeta} (\mathbf{s}) :=        \label{eq:phi_x}\\
&\  \frac{1}{\sqrt{D}}\!\left[\sin(\bbzeta_1^\top \mathbf{s})~ \cos(\bbzeta_1^\top \mathbf{s})~ \cdots~ \sin(\bbzeta_D^\top \mathbf{s})~ \cos(\bbzeta_D^\top \mathbf{s})\right]^\top\!\!. \nonumber
\end{align}
Consider now the RF-based linear function approximant
\begin{equation}
{\check v} (\mathbf{s}) =  \bbphi_{\zeta}^\top (\mathbf{s}) \bbtheta,~{\rm with}~ p(\bbtheta) = \mathcal{N}(\bbtheta; \mathbf{0}_{2D}, \sigma_\theta^2\mathbf{I}_{2D})  \label{eq:f_check}
\end{equation}
where $\sigma_\theta^2:=\kappa (0) = \sigma_\theta^2\bar{\kappa}(0)$. The prior pdf of $\check{\mathbf{v}}_t:=[\check{v}(\mathbf{s}_1)~ \check{v}(\mathbf{s}_2) ~\cdots$ $ \check{v}(\mathbf{s}_t)]^\top$ then becomes
\begin{equation}
p(\check{\mathbf{v}}_t|\mathbf{S}_t) = \mathcal{N} (\check{\mathbf{v}}_t; \mathbf{0}_t,
\check{\mathbf{K}}_t),~{\rm with}~\,  \check{\mathbf{K}}_t = \sigma_\theta^2\bbPhi_t \bbPhi_t^{\top} \label{eq:AGP}
\end{equation}
where $\bbPhi_t:=[\bbphi_{\zeta}(\mathbf{s}_1)~ \bbphi_{\zeta}(\mathbf{s}_2)~ \cdots~ \bbphi_{\zeta}(\mathbf{s}_t) ]^\top$. Evidently, $\check{\mathbf{K}}_t = \sigma_\theta^2\bbPhi_t \bbPhi_t^{\top}$ is a low-rank approximant of $\mathbf{K}_t$ in \eqref{eq:GP_Prior} when $t >2D$. With this approximate GP prior, the parametric function ${\check v}$ enjoys the nonparametric GP model expressiveness, with the simplicity of linear models. 

Specifically for GPTD, this RF-based GP prior \eqref{eq:AGP} allows us to approximate the predictor in \eqref{eq:batch_estimator} through
 \begin{subequations}
	\begin{align}
	\vspace*{-0.4cm}
&	\hat{\check{v}}(\mathbf{s}) \!= \! \bbphi_{\zeta}^\top\!(\mathbf{s}) \bigg(\! \bbPhi_{t+1}^{\top}\mathbf{H}_t^{\top}\mathbf{H}_t  \bbPhi_{t+1}\!+\! \frac{\sigma_n^2}{\sigma_\theta^2}\mathbf{I}_{2D}
	\bigg)^{-1}\!\!\!\!\!\! \bbPhi_ {t+1}^\top\mathbf{H}_t^{\top} \mathbf{r}_t \label{eq:mean_GP_RF}
		\vspace{-.4cm}\\
&
	\sigma_{\check{v}}^2 (\mathbf{s}) \!= \!
\bbphi_{\zeta}^\top\!(\mathbf{s}) \bigg(\!\frac{\bbPhi_{t+1}^{\top}\mathbf{H}_t^{\top}\!\mathbf{H}_t  \bbPhi_{t+1}}{\sigma_n^2} \!
+\! \frac{\mathbf{I}_{2D}}{\sigma_\theta^2} \bigg)^{-1}\!\!\!\!\!\!\!\bbphi_{\zeta}\!(\mathbf{s})  \label{eq:var_GP_RF}
	\end{align}	
\end{subequations}
which incurs computational complexity $\mathcal{O}(2Dt(t+1)+(2D)^3)$, that is certainly more affordable than the plain-vanilla GP predictor in \eqref{eq:mean_GP} and \eqref{eq:var_GP}, for  $t\ge 2D$.

Besides being scalable, the parametric function approximant $\check{v}$ readily accommodates streaming data $\mathbf{s}_{t+1}$ and $r_t$ in a recursive Bayesian fashion, leading to our online scalable (OS-) GPTD approach. Upon replacing $v$ by $\check{v}$ in the observation model \eqref{eq: r_t}, the per datum likelihood is 
\begin{equation}
p(r_{t}|\bbtheta, \mathbf{S}_{t+1}) = \mathcal{N}(r_{t}; \mathbf{h}_{t}^\top \bbtheta, \sigma_n^2) \label{eq:SGP_LL}
\end{equation}
where $\mathbf{h}_{t} := \bbphi_{\zeta} (\mathbf{s}_{t}) - \gamma\bbphi_{\zeta} (\mathbf{s}_{t+1})$. 
This linear Gaussian likelihood of \eqref{eq:SGP_LL} then propagates the Gaussianity from the prior (cf. \eqref{eq:f_check}) to the posterior across time. 

Capitalizing on the past rewards $\mathbf{r}_{t-1}$ and states $\mathbf{S}_t$ up until slot $t-1$, the agent keeps track of the posterior pdf of $\bbtheta$ as $p(\bbtheta|\mathbf{r}_{t-1},\mathbf{S}_{t}) = \mathcal{N}(\bbtheta; \hat{\bbtheta}_{t-1},\bbSig_{t-1})$, where $\hat{\bbtheta}_{t-1}$ and $\bbSig_{t-1}$ are the posterior mean and the posterior 
covariance matrix up until slot $t-1$, respectively; and with initial $\bm{\theta}_0 = \mathbf{0}_{2D}$ and $\bm{\Sigma}_0=\sigma_{\theta}^2\mathbf{I}_{2D}$ based on \eqref{eq:f_check}. 
Before receiving future rewards $\{r_t, r_{t+1}, \ldots\}$ that determine the value function $\check{v}(\cdot)$ at $\mathbf{s}_t$, the agent predicts the pdf of $\check{v}_t:=\check{v}(\mathbf{s}_t)$ using
\begin{equation}
p(\check{v}_t | \mathbf{r}_{t-1}, \mathbf{S}_t) = \mathcal{N}(\check{v}_t; \hat{\check{v}}_{t|t-1}, \sigma_{t|t-1}^2) \nonumber
\end{equation}
where the predictive mean and variance are
\begin{subequations}  	\label{eq:FA_SGP}
	\begin{align}
	\hat{\check{v}}_{t|t-1} & = \bbphi_{\zeta}^\top(\mathbf{s}_t) \hat{\bbtheta}_{t-1}  \label{eq:pred_mean}\\
	\sigma_{t|t-1}^2  & = \bbphi_{\zeta}^\top(\mathbf{s}_t) \bbSig_{t-1} \bbphi_{\zeta}(\mathbf{s}_t)  . \label{eq:pred_var}
	\end{align}
\end{subequations}

Upon taking action $a_t \in \mathcal{A}$ by following given policy $\mu$ at the beginning of time slot $t$, the agent receives reward $r_t$ and observes the new state of the environment $\mathbf{s}_{t+1}$, based on which the posterior pdf of $\bbtheta$ can be found by invoking the Bayes' rule as 
\begin{align}
p(\bbtheta|\mathbf{r}_{t}, \mathbf{S}_{t+1})& = \frac{p(\bbtheta|\mathbf{r}_{t-1},\mathbf{S}_{t}) p(r_{t}|\bbtheta, \mathbf{S}_{t+1})}{p(r_{t}|\mathbf{r}_{t-1})}  \nonumber\\
&= \mathcal{N}(\bbtheta; \hat{\bbtheta}_{t}, \bbSig_{t}) \nonumber
\end{align}
where the moments are recursively updated as follows
\begin{subequations} \label{eq:SGP_up}
\begin{align}
\hat{\bbtheta}_{t} &= \hat{\bbtheta}_{t-1} +
\frac{\bbSig_{t-1} \mathbf{h}_{t}(r_{t} - \mathbf{h}_{t}^\top \hat{\bbtheta}_{t-1})}{\mathbf{h}_{t}^\top\bbSig_{t-1}\mathbf{h}_{t}+ \sigma_n^2 }   \label{eq:mean_up} \\
\bbSig_{t} &= \bbSig_{t-1} -  \frac{ \bbSig_{t-1} \mathbf{h}_{t}\mathbf{h}_{t}^\top\bbSig_{t-1}}{\mathbf{h}_{t}^\top\bbSig_{t-1}\mathbf{h}_{t}+ \sigma_n^2}  \label{eq:cov_up}.
\end{align}
\end{subequations}

Besides offering a means of quantifying the uncertainty of the estimate in \eqref{eq:mean_up}, the covariance matrix \eqref{eq:cov_up} also adjusts the learning rate adaptively from a stochastic optimization point of view. Moreover, it is evident that the computational overhead over $t$ slots is $\mathcal{O}(t(2D)^2)$ while the storage requirement is $\mathcal{O}((2D)^2)$. Hence, scalability is also guaranteed for our OS-GPTD scheme.

\section{Worst-case performance analysis}\label{sec:perform}

The value function estimator along with its variance in \eqref{eq:FA_SGP} were derived based on the assumed generative model \eqref{eq: r_t} and a GP prior on $v(\cdot)$. One would be interested in benchmarking the performance of the sequence of estimates \eqref{eq:mean_up} even if such modeling assumptions are violated. Considering the worst-case scenario in which reward-state pairs are chosen by an adversary, performance is commonly evaluated in terms of the so-called \textit{regret}.
Regret analysis has well-documented merits in online learning and optimization \cite{hazan2007logarithmic}.
For a given loss, it benchmarks the performance of an online learner (a.k.a., algorithm) against the best fixed strategy that can utilize data in hindsight. In our online TD learning context, the loss can be simply selected as Bellman's error. Beyond establishing logarithmic regret on this error, we can further bound the long-term reward prediction error of OS-GPTD relative to the benchmark value function. 

Worst-case Bellman's error and the long-term reward prediction error will be analyzed in the 
rest of this section. It is worth stressing that for a given trajectory of state-reward pairs, our analyses pertain to the sequence of parameter estimates produced by \eqref{eq:SGP_up}, without any probabilistic assumptions stated in the previous section.

\subsection{Bellman error regret analysis}
Regret analysis of our OS-GPTD algorithm will be quantified first in terms of Bellman's error, which demonstrates the self-consistency of the value function evaluated at two successive states. For the RF-based OS-GPTD, the instantaneous Bellman error at slot $t$ for any $\bbtheta$ is simply $l_t(\bbtheta):=(r_t - \mathbf{h}_t^\top \bbtheta)^2$ \cite{sutton2018reinforcement,sun2015online}. With batch data in hindsight, the benchmark function $v^{*}$ belongs to the $\kappa$-induced reproducing kernel Hilbert space (RKHS), as we have seen in \eqref{eq:mean_GP}. The Bellman error functional of $v^\ast$ is thus $l_t (v^*):=(r_t - v^{*} (\mathbf{s}_{t})+\gamma v^{*} (\mathbf{s}_{t+1}))^2$. 
Hence, the cumulative regret of OS-GPTD over $T$ slots is
\begin{equation}\label{eq:regret_1}
\mathcal{R}(T): = \sum_{t=1}^T l_t (\hat{\bbtheta}_{t-1}) -  \sum_{t=1}^T l_t (v^{*}).
\end{equation}
Let $\bbtheta^{*}$ denote the best fixed RF-based parameter vector given the entire state-reward trajectory. The regret in \eqref{eq:regret_1} can then be readily decomposed as
\begin{equation}
\mathcal{R}(T) = \underbrace{\sum_{t=1}^T\! l_t (\hat{\bbtheta}_{t-1}) \!-\! \!\sum_{t=1}^T\! l_t (\bbtheta^{*})\!}_{:=\mathcal{R}_1 (T)} +\! \underbrace{\sum_{t=1}^T\! l_t (\bbtheta^{*})\! -\!\!  \sum_{t=1}^T\! l_t( v^*)}_{:=\mathcal{R}_2 (T)}   \nonumber
\end{equation}
where 
$\mathcal{R}_1 (T)$ corresponds to the regret incurred by confining ourselves to learn within the class of RF-induced parametric functions, and $\mathcal{R}_2 (T)$ captures the accumulated RF-based approximation error of the kernel. To upper-bound $\mathcal{R}_1 (T)$ and $\mathcal{R}_2 (T)$, we will start with two standard assumptions.
\begin{itemize}
	\item[{\bf as1)}] The Bellman error is bounded by $B_e$; and,
		\item[{\bf as2)}] The kernel $\bar{\kappa}$ is shift-invariant, standardized, and bounded such that $\bar{\kappa} (\mathbf{s}, \mathbf{s}') \leq 1$, $ \forall \mathbf{s}, \mathbf{s}'$.
\end{itemize}

Boundedness under as1) is always satisfied for bounded rewards and state-conditioned features, when the RF-based parameter vector or the RKHS-based function is bounded as well.
On the other hand, conditions in {as2)} hold for a broad family of kernels including e.g., Gaussian, Laplace, and Cauchy ones~\cite{rahimi2008random}.

The ensuing lemma upper bounds the regret $\mathcal{R}_1 (T)$, whose proof is relegated to Appendix \ref{pf:le} of the supplementary material.
\begin{lemma}\label{le:r1}
Under as1), the following bound holds for the cumulative Bellman error incurred by the OS-GPTD relative to its counterpart by the best fixed parameter vector $\bbtheta^{*}$ in hindsight
	\begin{equation}
\mathcal{R}_1(T)\leq 2D B_e^2 \log\! \left(\frac{(1+\gamma)^2\sigma_{\theta}^2}{\sigma_n^2} T +1 \right) + \frac{\| \bbtheta^{*}\|_2^2 }{2 \sigma_{\theta}^2}.
	\end{equation}
\end{lemma}


Lemma \ref{le:r1} establishes the regret bound concerning the cumulative Bellman error from the online RF-based function estimator relative to that from the best fixed one with data in hindsight. In the next theorem, we further account for the difference of the Bellman errors arising from approximating the RKHS-based function estimator with the RF-based linear one. This will lead us to a bound for
the overall regret $\mathcal{R}(T)$ in \eqref{eq:regret_1}.

\begin{theorem}\label{th:1}
Under {as1)}-{as2)} and with  $v^{*}$ belonging to the RKHS $\mathcal{H}$ induced by $\kappa$, for any fixed $\epsilon >0$, the following bound holds with probability at least $1- \frac{2^8\sigma_{\zeta}^2}{\epsilon^2} \,{ e}^{-\frac{D\epsilon^2}{4d+8} }$
\begin{align}
&\sum_{t = 1}^T l_t (\hat{\bbtheta}_{t-1})\!-\!  \sum_{t = 1}^T l_t (v^{*}) \leq 2(1+\gamma)\epsilon B_e C T   \nonumber\\
&~ + \frac{(1+\epsilon)C^2}{2\sigma_{\theta}^2}+  2D B_e^2 \log\! \left(\frac{(1+\gamma)^2\sigma_{\theta}^2}{\sigma_n^2}  T +1 \right) \label{eq:theorem1}
\end{align}
where $C>0$ is some constant, and $\sigma_{\zeta}^2 : = \mathbb{E}_{\pi_{\bar{\kappa}}}[\| \bbzeta\|^2]$ is the second-order moment of the RF vector $\bbzeta$. 
Setting $\epsilon = \mathcal{O}(\log T/T)$, the regret in \eqref{eq:theorem1} boils down to 
\begin{equation}
\mathcal{R}(T) = \mathcal{O}(\log T). \label{eq:reg_full}
\end{equation}
\end{theorem}


With the total regret over $T$ slots being logarithmic in $T$, our proposed OS-GPTD algorithm achieves \emph{no regret} on average in terms of the Bellman error.

\subsection{Long-term reward prediction error bound}
For a given state-reward trajectory, the empirical value function or the discounted sum of future rewards at slot $t$ is $\bar{v}_t := \sum_{\tau=t}^{\infty} \gamma^{\tau-t} r_{\tau}$. OS-GPTD then incurs online prediction error of the long-term reward given by $e_t := \hat{\check{v}}_{t|t-1} - \bar{v}_t $, where $\hat{\check{v}}_{t|t-1} $ is the discounted future reward predicted as in \eqref{eq:pred_mean} based on the historical rewards up to slot $t-1$. Although the no-regret property of OS-GPTD in the Bellman error demonstrates that the estimated value function is self-consistent at any two consecutive states, it does not necessarily imply small long-term reward prediction errors, as it is also noted by \cite{sun2015online}. To further benchmark the performance of OS-GPTD regarding its long-term reward prediction error, we will compare the cumulative online prediction error $\sum_{t=1}^T e_t$ with its counterpart from the best fixed function $v^{*}$, namely, $\sum_{t=1}^T e_t^{*}$ with $e_t^{*} := v^{*}(\mathbf{s}_t) - \bar{v}_t$. While being not necessary, the following performance bound conveniently builds upon the no-regret property established in the previous subsection.
%
%

\begin{theorem}\label{th:2}
Let as1)-as2) hold, and consider the sequence of RF-based
linear 
value function estimators produced recursively by \eqref{eq:mean_up}.
The next bound holds w.h.p. on the average online prediction error of the long-term rewards relative to its counterpart from the best fixed 
estimator $v^{*}$, which belongs to the $\kappa$-induced RKHS $\mathcal{H}$ 
\begin{equation}
\lim_{T\rightarrow \infty} \frac{\sum_{t=1}^T  e_t^2}{T}   \leq \frac{2(1+\gamma)^2}{(1-\gamma)^2} \frac{\sum_{t=1}^T  e_t^{*2}}{T}  . \label{eq:PE_bound}
\end{equation}
\end{theorem}

\section{Scalable TD learning with ensemble GPs}
While a scalable approach has been devised in Section \ref{sec:Single_GP} to estimate the state value function on-the-fly, its performance hinges largely on the \emph{preselected} kernel of the GP prior, which confines expressiveness of the function space. Aiming at constructing a richer function space\cite{lu2020ensemble,lu2021incremental}, we advocate in this section an \emph{ensemble} of $M$ GPs (experts) for the agent, each postulating a unique GP prior on $v$, denoted as $v|m \sim \mathcal{GP}(0, \kappa^m (\mathbf{s}, \mathbf{s}'))$, where $m\in \mathcal{M}:=\{1,\ldots,M\}$ is the expert index and $\kappa^m$ is a shift-invariant kernel selected from a {\it prescribed} dictionary $\mathcal{K}:=\{\kappa^1, \ldots, \kappa^M\}$. Per expert $m$, the prior pdf of the value function $v$ evaluated at $\mathbf{S}_t$ is 
\begin{equation}
p(\mathbf{v}_t|z=m, \mathbf{S}_t) \!=\! \mathcal{N} (\mathbf{v}_t ; {\bf 0}_t, {\bf K}^m_t ) \nonumber 
\end{equation}
where the hidden association variable $z \in \mathcal{M}$ is introduced to indicate the existence of each expert in the ensemble (E) GP, and ${\bf K}_t^m$ is the covariance matrix induced by $\kappa^m$. Considering the $M$ GP experts jointly, the EGP priors on $\mathbf{v}_t$ from the Gaussian mixture (GM) given by
\begin{equation}
p(\mathbf{v}_t|\mathbf{S}_t) = \sum_{m = 1}^M w^m \mathcal{N}(\mathbf{v}_t; {\bf 0}_t, \mathbf{K}^m_t) \label{eq:EGP_prior} 
\end{equation}
where the weights $\{w^m\}_{m=1}^M$ are viewed as the prior probabilities of each GP expert to be present in the EGPs. 

As with a single GP prior or expert, the computational complexity of exact function estimation with the EGP prior \eqref{eq:EGP_prior} will be intractable when the number of the state-rewards grows large. Again, we will leverage RF-based kernel approximants for the value function to effect scalability in the EGP prior, contributing to our novel OS-EGPTD approach.

Seeking a tractable function predictor, each GP expert $m$ resorts to the $\kappa^m$ induced RF-based approximant, where $\kappa^m = \sigma_{\theta^m}^2\bar{\kappa}^m$. Upon drawing vectors $\{\bbzeta_i^m\}_{i=1}^D$ i.i.d. from $\pi_{\bar{\kappa}}^m (\bbzeta)$,  expert $m$ relies on the RF vector \eqref{eq:phi_x} and  the per-expert parameter vector $\bbtheta^m$ to yield the generative model for the linear approximant $\check{v}$ 
\begin{subequations}
	\begin{align}
	{p}(\bbtheta^m|z=m) &= \mathcal{N} (\bbtheta^m; \mathbf{0}_{2D}, \sigma_{\theta^m}^2\mathbf{I}_{2D}) \label{eq:p_theta}\\
	p(\check{v}(\mathbf{s})|\bbtheta^m ,z=m)& = \delta\big(\check{v}(\mathbf{s}) -  \bbphi_{\zeta}^{m\top} (\mathbf{s})\bbtheta^m\big) \label{eq:f_s} 
	\end{align}
\end{subequations}
where $\delta$ is Kronecker's delta. This linear approximant implies that the data likelihood at slot $t$ for expert $m$ is 
\begin{equation}
p(r_t| \bbtheta^m, \mathbf{S}_{t+1}, z = m) = \mathcal{N}\big(r_t; \mathbf{h}_t^{m\top}\bbtheta^m, \sigma_n^2\big) \label{eq:LL}
\end{equation}
where the per-expert $m$ observation vector is
\begin{equation}
\mathbf{h}_t^{m} := \bbphi^m_{\zeta}(\mathbf{s}_{t}) - \gamma \bbphi^m_{\zeta}(\mathbf{s}_{t+1}). \label{eq:h_t}
\end{equation}
Aided by the Gaussian prior for $\bbtheta^m$ (cf. \eqref{eq:p_theta}), this linear Gaussian likelihood \eqref{eq:LL} will lead to a Gaussian posterior $p(\bbtheta^m |\mathbf{r}_{t}, \mathbf{S}_{t+1}, z=m) = \mathcal{N}(\bbtheta^m; \hat{\bbtheta}_{t}^m, \bbSig^m_{t})$ characterized by its mean $\hat{\bbtheta}_{t}^m$ and covariance matrix $\bbSig^m_{t}$. The agent will rely on  $w_t^m := {\rm Pr}(z = m| \mathbf{r}_t, \mathbf{S}_{t+1})$ to account for the contribution of expert $m$. With sequential pairs $(r_t, \mathbf{s}_{t+1})$ collected by the agent, the $M$ triplets $\{w_{t}^m, \hat{\bbtheta}_{t}^m, \bbSig^m_{t}\}_{m=1}^M$ summarizing the past information about the state value function, will be updated by alternating between prediction and correction steps as follows.

\subsection{Prediction}

Upon collecting $\{w_{t-1}^m, \hat{\bbtheta}_{t-1}^m, \bbSig^m_{t-1}\}_{m=1}^M$ from the $M$ GP experts at the end of slot $t-1$, the agent predicts the per-expert pdf of $v(\mathbf{s}_t)$ through 
\begin{align}
&p(\check{v}_t| z \!=\! m, \mathbf{r}_{t-1}, \mathbf{S}_t) \!= \!\!\int\!\! p(\check{v}_t, \bbtheta^m| z \!=\! m, \mathbf{r}_{t-1}, \mathbf{S}_t) d\bbtheta^m \nonumber\\
& = \!\int\! \delta\big(\check{v}_t -  \bbphi_{\zeta}^{m\top} (\mathbf{s}_t)\bbtheta^m\big)p(\bbtheta^m| \mathbf{r}_{t-1}, \mathbf{S}_t) d\bbtheta^m \nonumber\\
& = \mathcal{N}\big(\check{v}_t; \hat{\check{v}}_{t|t-1}^m, (\sigma^{m}_{t|t-1})^2\big)
\end{align}
where the predictive mean and variance are \vspace{-.2cm}
\begin{subequations} 
\label{eq:EGP_single}
	\begin{align}
	\hat{\check{v}}_{t|t-1}^m & = \bbphi_{\zeta}^{m\top}(\mathbf{s}_t) \hat{\bbtheta}_{t-1}^m  \\
	(\sigma^{m}_{t|t-1})^2 & = \bbphi_{\zeta}^{m\top}(\mathbf{s}_t) \bbSig^m_{t-1} \bbphi_{\zeta}^m(\mathbf{s}_t)  .
	\end{align}
\end{subequations}
The sum and product probability rules further enable the agent to obtain the ensemble predictive pdf of $v(\mathbf{s}_{t})$ as 
\begin{align}
&p(\check{v}(\mathbf{s}_t)| \mathbf{r}_{t-1}, \mathbf{S}_t ) = \sum_{m=1}^M p(\check{v}(\mathbf{s}_t), z = m| \mathbf{r}_{t-1}, \mathbf{S}_t) \nonumber\\
& = \sum_{m=1}^M w_{t-1}^m \mathcal{N}\big(\check{v}(\mathbf{s}_t); \hat{\check{v}}_{t|t-1}^m, (\sigma^{m}_{t|t-1})^2\big)
\end{align}
which is a weighted superposition of the predictive pdfs from all experts. The minimum mean-square error 
predictor of $v(\mathbf{s}_t)$ can be obtained along with its variance as \vspace{-.2cm}
\begin{subequations}\label{eq:v_pre_static}
	\begin{align}
\hspace*{-0.2cm}
\bar{\check{v}}_{t|t-1} \! & =  \! \sum_{m = 1}^M w_{t-1}^m \hat{\check{v}}_{t|t-1}^m \label{eq:mean_v_pre} \\	
\hspace*{-0.2cm}	\bar{\sigma}_{t|t-1}^2\!  & =\! \sum_{s = 1}^m w_{t-1}^m \big[(\sigma^{m}_{t|t-1})^2 \!+\!  (\bar{\check{v}}_{t|t-1}\! -\! \hat{\check{v}}_{t|t-1}^m)^2\big]. \label{eq:var_v_pre}
	\end{align}
	\vspace{-.2cm}
\end{subequations} 

\begin{algorithm}[t]
	\caption{OS-EGPTD}
	\label{alg:EGPTDL}
	\begin{algorithmic}
		\State{\textbf{Input:}  $\kappa^m$, $\sigma_{\theta^m}^2$, $m = 1,\ldots, M$, $\sigma_n^2$, $D$, $\gamma$ and $\mathbf{s}_1$}.
	    \State{\textbf{Initialization:}  }
		\For{m = 1, 2, \ldots, M}
		\State Draw $D$ RF vectors $\{\bbzeta_i^m\}_{i = 1}^D$;
		\State  $w_0^m = 1/M$;	$\hat{\bbtheta}_0^m = \mathbf{0}_{2D}$; $\bbSig_{0}^m = \sigma_{\theta^m}^2\mathbf{I}_{2D}$;
		\EndFor
		\For{$t = 1, 2, \ldots, T$}
		\State Find predictive mean/variance of $\check{v}(\mathbf{s}_t)$ via \eqref{eq:v_pre_static};
		\State Receive $(r_t, \mathbf{s}_{t+1})$;
		\For{$m = 1, 2, \ldots, M$} 
		\State Construct RF $\bbphi_{\zeta}^m(\mathbf{s}_{t+1})$ via \eqref{eq:phi_x};
		\State Construct the observation vector $\mathbf{h}^m_t$ via \eqref{eq:h_t};
	    \State  \begin{varwidth}[t]{\linewidth}
			Update per-expert pdf of $\bbtheta^m$ via \eqref{eq:p_theta_up_1};
		\end{varwidth}	 
		\State Update $w_{t}^m$ via \eqref{eq:w_update_1};
		\EndFor
		\EndFor
	\end{algorithmic}
\end{algorithm}

\vspace{-.2cm}
\subsection{Correction}

After receiving $r_t$ and $\mathbf{s}_{t+1}$ at slot $t$, GP expert $m$ employs the data likelihood \eqref{eq:LL} to propagate the posterior pdf as 
\begin{align}
&{p}(\bbtheta^m |\mathbf{r}_{t}, \mathbf{S}_{t+1}, z=m) \label{eq:p_theta_up_1} \\
&= 	\frac{{p}(\bbtheta^m |\mathbf{r}_{t-1}, \mathbf{S}_{t}, z=m) {p}(r_{t}|\bbtheta^m, 
	\mathbf{S}_{t},z=m)}{{p}(r_{t}|\mathbf{r}_{t-1}, \mathbf{S}_{t+1}, z=m)} \nonumber \\
& = \mathcal{N}(\bbtheta^m; \hat{\bbtheta}_{t}^m, \bbSig^m_{t}) \nonumber
\end{align}
where the mean and covariance matrix are updated using 
\begin{subequations} \label{eq:posterior_update_1}
	\begin{align}
	\hat{\bbtheta}_{t}^m &= \hat{\bbtheta}_{t-1}^m +
	\frac{\bbSig_{t-1}^m \mathbf{h}_{t}^m\big(r_{t} - \mathbf{h}_{t}^{m\top} \hat{\bbtheta}^m_{t-1}\big)}{\mathbf{h}_{t}^{m\top}\bbSig_{t-1}^m\mathbf{h}_{t}^m + \sigma_n^2 }   \label{eq:mean_up_2} \\
	\bbSig_{t}^m &= \bbSig_{t-1}^m -\frac{ \bbSig_{t-1}^m \mathbf{h}_{t}^m\mathbf{h}_{t}^{m\top}\bbSig_{t-1}^m}{\mathbf{h}_{t}^{m\top}\bbSig_{t-1}^m\mathbf{h}_{t}^m+ \sigma_n^2}  \label{eq:cov_up_2} .
	\end{align}
\end{subequations}
Meanwhile, the per-expert posterior weight is corrected using Bayes' rule as  
\begin{align}
&w_{t}^m = {\rm Pr}(z=m|\mathbf{r}_{t}, \mathbf{S}_{t+1}) \nonumber\\
&= \frac{{\rm Pr}(z=m|\mathbf{r}_{t-1}, \mathbf{S}_{t}) {p}(r_{t}|\mathbf{r}_{t-1}, \mathbf{S}_{t+1},  z=m)}{\sum_{m=1}^M {\rm Pr}(z=m|\mathbf{r}_{t-1}, \mathbf{S}_{t}) {p}(r_{t}|\mathbf{r}_{t-1}, \mathbf{S}_{t+1},  z=m)} \nonumber
\end{align}
which, upon utilizing 
\begin{align}
&{p}(r_{t}|\mathbf{r}_{t-1}, \mathbf{S}_{t+1},  z=m) \nonumber\\
&= \int\! {p}(r_{t}|\bbtheta^m, \mathbf{S}_{t+1},  z=m) {p}(\bbtheta^m|\mathbf{r}_{t-1}, \mathbf{S}_{t},  z=m) d \bbtheta^m \nonumber\\
&=\mathcal{N}\big (r_{t}; \mathbf{h}_t^{m\top}\hat{\bbtheta}^m_{t-1}, \mathbf{h}_t^{m\top}\bbSig^m_{t-1}\mathbf{h}_t^{m}+\sigma_n^2\big) \nonumber
\end{align}
boils down to  
\begin{align}
&w_{t}^m  =  \label{eq:w_update_1} \\
&\frac{w_{t-1}^m \mathcal{N} \big(r_{t}; \mathbf{h}_t^{m\top}\hat{\bbtheta}^m_{t-1}, \mathbf{h}_t^{m\top}\bbSig^m_{t-1}\mathbf{h}_t^{m}+\sigma_n^2\big)}{\sum_{m' = 1}^M w_{t-1}^{m'} \mathcal{N}\big (r_{t}; \mathbf{h}_t^{m'\top}\hat{\bbtheta}^{m'}_{t-1}, \mathbf{h}_t^{m'\top}\bbSig^{m'}_{t-1}\mathbf{h}_t^{m'}\!+\!\sigma_n^2\big)}.   \nonumber
\end{align}
Accounting for all $M$-expert updates, the novel OS-EGPTD scheme, summarized in Algorithm 1, runs $M$ OS-GPTDs in parallel, thus incurring a per-slot complexity of $\mathcal{O}(M (2D)^2)$. Hence, scalability is not compromised by the ensemble approach that also endows the value function $v$ with richer expressiveness.

\begin{remark}
Although the focus so far has been on the state value function estimation, our OS-(E)GPTD methods can be readily employed to estimate state-action value functions, thus accommodating policy improvement without any additional knowledge on the environment or MDP.
\end{remark}

%
%
%
%
%
%
%

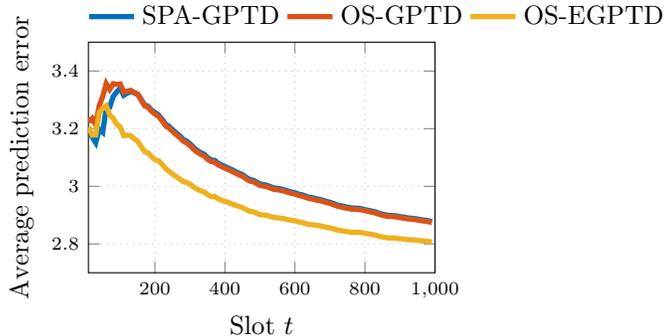
\begin{figure}[t]
	\centering{
%
%
\definecolor{mycolor1}{rgb}{0.00000,0.44700,0.74100}%
\definecolor{mycolor2}{rgb}{0.85000,0.32500,0.09800}%
\definecolor{mycolor3}{rgb}{0.92900,0.69400,0.12500}%
\begin{tikzpicture}

\begin{axis}[%
width=0.95\mywidth,
height=1.1\myheight,
at={(1.011in,0.642in)},
scale only axis,
xmin=10,
xmax=1000,
xlabel style={font=\xlabelfontsize},
xlabel={Slot $t$},
ymin=2.7,
ymax=3.5,
ylabel style={font=\ylabelfontsize},
ylabel={Average prediction error},
axis background/.style={fill=white},
xmajorgrids,
ymajorgrids,
grid style={dotted},ticklabel style={font=\ticklabelfontsize},
legend columns=3,
legend style={
	at={(-0.03,1.015)}, 
	anchor=south west, legend cell align=left, align=left, draw=none
	, font=\legendfontsize}
]
\addplot [color=mycolor1,line width=2.0pt]
  table[row sep=crcr]{%
11	3.2026813154942\\
21	3.17005882061666\\
31	3.15119605466707\\
41	3.19753216966523\\
51	3.18712600414041\\
61	3.272417258674\\
71	3.27477263857895\\
81	3.31220385734264\\
91	3.32773679631928\\
101	3.33996116015908\\
111	3.3156501365133\\
121	3.32334568789572\\
131	3.32948403762968\\
141	3.32378948280834\\
151	3.31973084706116\\
161	3.30073976466318\\
171	3.28145592169975\\
181	3.27713762923792\\
191	3.26358002734621\\
201	3.25506652887188\\
211	3.24680002848061\\
221	3.23148931352907\\
231	3.21481171143377\\
241	3.20739713656021\\
251	3.1955120140161\\
261	3.18478911094408\\
271	3.17473940552437\\
281	3.16257634708774\\
291	3.15632256230846\\
301	3.1466025053779\\
311	3.13384418522054\\
321	3.12415256978551\\
331	3.116730521211\\
341	3.11085786365877\\
351	3.09867308438183\\
361	3.0912004116145\\
371	3.08983810093926\\
381	3.07926761901455\\
391	3.07371263826758\\
401	3.06789885202612\\
411	3.0624344533599\\
421	3.05691337645259\\
431	3.050529249944\\
441	3.0448769299797\\
451	3.03961575098434\\
461	3.03064537328368\\
471	3.02337215363182\\
481	3.02088222273967\\
491	3.01429180853611\\
501	3.00761791796587\\
511	3.00567387036901\\
521	3.00292633228532\\
531	2.99804233778032\\
541	2.99375786550213\\
551	2.99280215993429\\
561	2.9903981097272\\
571	2.98720497607918\\
581	2.9830629070836\\
591	2.97973232172774\\
601	2.97734326064337\\
611	2.97267059102171\\
621	2.9700571170399\\
631	2.96580029018547\\
641	2.96171415585428\\
651	2.95994637535761\\
661	2.95638384283181\\
671	2.9544710775566\\
681	2.9507792169855\\
691	2.94720438625065\\
701	2.94415359716742\\
711	2.93977798998349\\
721	2.93494476466049\\
731	2.9325195745369\\
741	2.92947437721492\\
751	2.92737129563035\\
761	2.92474784305819\\
771	2.92418427926819\\
781	2.92365351768169\\
791	2.92187371976788\\
801	2.91864878267638\\
811	2.91603834285886\\
821	2.91344783889155\\
831	2.91056525966539\\
841	2.90614862312357\\
851	2.90291323269681\\
861	2.89998981063566\\
871	2.8985934779124\\
881	2.89845077867571\\
891	2.89697833598398\\
901	2.89443466934373\\
911	2.89304310654613\\
921	2.89046101018142\\
931	2.88940529985168\\
941	2.88803946099238\\
951	2.8860557185874\\
961	2.88392528314943\\
971	2.88233646800098\\
981	2.88048393005959\\
991	2.87756199676582\\
};
\addlegendentry{SPA-GPTD}

\addplot [color=mycolor2,line width=2.0pt]
  table[row sep=crcr]{%
11	3.22262576039421\\
21	3.23542788009797\\
31	3.21403944115751\\
41	3.27982606226976\\
51	3.31177736652643\\
61	3.35636678401612\\
71	3.33627675586703\\
81	3.35563287780327\\
91	3.353554397298\\
101	3.3547424659578\\
111	3.32614585956612\\
121	3.32973960995228\\
131	3.33289697996873\\
141	3.32506258239104\\
151	3.31906204851555\\
161	3.29878330071037\\
171	3.27880210997432\\
181	3.27363326862429\\
191	3.25929241910029\\
201	3.25017998088741\\
211	3.24176474728485\\
221	3.22615706384319\\
231	3.20927315962037\\
241	3.20174830198058\\
251	3.18981458807259\\
261	3.17902145884177\\
271	3.1688596490932\\
281	3.15658288275584\\
291	3.15033095052476\\
301	3.14060673073711\\
311	3.12783441616246\\
321	3.11820932363732\\
331	3.1108201513703\\
341	3.10497591297323\\
351	3.09284082121399\\
361	3.08540473853639\\
371	3.08409552535024\\
381	3.07362679083481\\
391	3.06808891439252\\
401	3.06233109841456\\
411	3.05690805074146\\
421	3.05144312123646\\
431	3.04511619947057\\
441	3.0395224952386\\
451	3.03427708239897\\
461	3.02535944269971\\
471	3.01816308377053\\
481	3.01572042780589\\
491	3.00918808343091\\
501	3.00257800628606\\
511	3.00069383723953\\
521	2.99802102178618\\
531	2.99319285170352\\
541	2.98895314526107\\
551	2.98803511106981\\
561	2.98567188901406\\
571	2.98252270800141\\
581	2.97842682341254\\
591	2.97512967239716\\
601	2.97280430096728\\
611	2.9681753740261\\
621	2.96560685644364\\
631	2.96139423838073\\
641	2.9573551280081\\
651	2.95561148651564\\
661	2.95209011254447\\
671	2.95022032468416\\
681	2.94656697650825\\
691	2.9430308174589\\
701	2.9400224585375\\
711	2.93569199332281\\
721	2.93089446177581\\
731	2.92850242545622\\
741	2.92549365778069\\
751	2.92342018284747\\
761	2.92082790904907\\
771	2.92029984785122\\
781	2.91980120814846\\
791	2.91806127597052\\
801	2.91487791176057\\
811	2.91230048668994\\
821	2.90974174709097\\
831	2.90688683411357\\
841	2.90250562658577\\
851	2.89930470068849\\
861	2.89641688247219\\
871	2.89505155913681\\
881	2.89493749017794\\
891	2.89349476940366\\
901	2.89098252897609\\
911	2.88961772125339\\
921	2.88706518718845\\
931	2.88604089166499\\
941	2.88470658532857\\
951	2.88274934976666\\
961	2.88064570267737\\
971	2.87908248660654\\
981	2.87725857737993\\
991	2.87436102642447\\
};
\addlegendentry{OS-GPTD}

\addplot [color=mycolor3, line width=2.0pt]
  table[row sep=crcr]{%
11	3.20598528387705\\
21	3.17779923543249\\
31	3.17838890113393\\
41	3.25572534642016\\
51	3.26967611927558\\
61	3.27927985644477\\
71	3.24648928763064\\
81	3.23595756823339\\
91	3.21459835814666\\
101	3.20584336340126\\
111	3.17541917301672\\
121	3.17708052645345\\
131	3.17512756217758\\
141	3.1646240854502\\
151	3.1556591722977\\
161	3.13742014590355\\
171	3.1197500306659\\
181	3.11546221147235\\
191	3.1020956258125\\
201	3.09176306179203\\
211	3.08827311212717\\
221	3.07516951450743\\
231	3.06108169749635\\
241	3.05161586772217\\
251	3.04146418271392\\
261	3.03351476892168\\
271	3.02467686600705\\
281	3.01635762552336\\
291	3.01293439654975\\
301	3.00682239790054\\
311	2.99700190834223\\
321	2.98986276594373\\
331	2.98424364245356\\
341	2.98018480266123\\
351	2.97033990486353\\
361	2.96463084565673\\
371	2.96448255843266\\
381	2.95644705663284\\
391	2.95125901156131\\
401	2.94733098503773\\
411	2.94306885318177\\
421	2.93815313242467\\
431	2.9336351160908\\
441	2.92987622762032\\
451	2.92595642613702\\
461	2.9188990930338\\
471	2.91336016467252\\
481	2.91143392085622\\
491	2.90650103247454\\
501	2.90127044905238\\
511	2.90068158175207\\
521	2.89880516805069\\
531	2.89463234159244\\
541	2.89125600813773\\
551	2.89062469269363\\
561	2.88922017120863\\
571	2.88700769510103\\
581	2.88425198113538\\
591	2.88158979982784\\
601	2.88008314160409\\
611	2.8762539159177\\
621	2.87404882628131\\
631	2.87050586910842\\
641	2.86736562019201\\
651	2.86658233881021\\
661	2.86443494377797\\
671	2.86300456320813\\
681	2.86011451394964\\
691	2.85747004454424\\
701	2.85546520955857\\
711	2.85211635430799\\
721	2.8481475632259\\
731	2.84625045957045\\
741	2.84418492129876\\
751	2.8423977523673\\
761	2.84027643058085\\
771	2.84010888282826\\
781	2.84044185894689\\
791	2.83917715721088\\
801	2.83651685429298\\
811	2.83450534230152\\
821	2.83249438931409\\
831	2.83038007995232\\
841	2.82679086724201\\
851	2.82387346504019\\
861	2.82157637718521\\
871	2.82056379348783\\
881	2.82091118789823\\
891	2.81979867947465\\
901	2.81783906862378\\
911	2.81700279393541\\
921	2.81490129308219\\
931	2.81446118145348\\
941	2.81380331751496\\
951	2.81238611652445\\
961	2.81083948146427\\
971	2.80979970486817\\
981	2.80870330859789\\
991	2.80632009409332\\
};
\addlegendentry{OS-EGPTD}

\end{axis}
\end{tikzpicture}%
}
	\caption{Average prediction error on {\bf Random Walk}.}
	\label{fig:RW_Ve}
\end{figure}

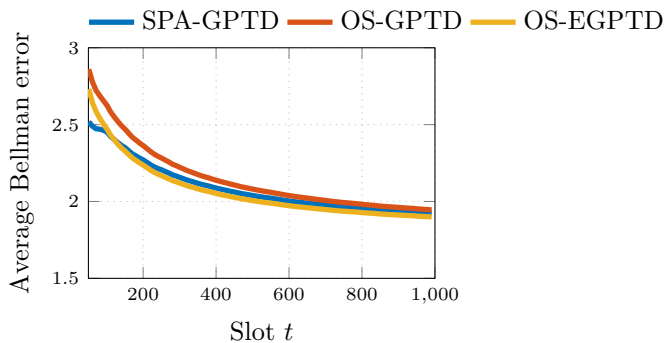
\begin{figure}[t]
	\centering{
%
%
\definecolor{mycolor1}{rgb}{0.00000,0.44700,0.74100}%
\definecolor{mycolor2}{rgb}{0.85000,0.32500,0.09800}%
\definecolor{mycolor3}{rgb}{0.92900,0.69400,0.12500}%
\begin{tikzpicture}

\begin{axis}[%
width=0.95\mywidth,
height=1.1\myheight,
at={(1.011in,0.642in)},
scale only axis,
xmin=50,
xmax=1000,
xlabel style={font=\xlabelfontsize},
xlabel={Slot $t$},
ymin=1.5,
ymax=3,
ylabel style={font=\ylabelfontsize},
ylabel={Average Bellman error},
axis background/.style={fill=white},
xmajorgrids,
ymajorgrids,
grid style={dotted},ticklabel style={font=\ticklabelfontsize},
legend columns=3,
legend style={
	at={(-0.03,1.015)}, 
	anchor=south west, legend cell align=left, align=left, draw=none
	, font=\legendfontsize}
]
\addplot [color=mycolor1, line width=2.0pt]
  table[row sep=crcr]{%
51	2.51822334041867\\
61	2.48988896635235\\
71	2.47335922978678\\
81	2.47016805179186\\
91	2.46333069420318\\
101	2.45187503775745\\
111	2.42279465541519\\
121	2.40233923979504\\
131	2.38245820350596\\
141	2.3641854615427\\
151	2.34948232025873\\
161	2.32880035301633\\
171	2.30876229463572\\
181	2.29590269591594\\
191	2.28121664069755\\
201	2.26905465337292\\
211	2.25364383518978\\
221	2.23765609428842\\
231	2.22454110070229\\
241	2.21406780375195\\
251	2.20541126778711\\
261	2.19495829819269\\
271	2.18469017457173\\
281	2.17255559731331\\
291	2.16539739026697\\
301	2.15634617849425\\
311	2.14750390545602\\
321	2.13910504782731\\
331	2.13209461963083\\
341	2.12493932114352\\
351	2.11732222203166\\
361	2.11084209359857\\
371	2.10520716776062\\
381	2.09980276559855\\
391	2.09280060798435\\
401	2.08692412571463\\
411	2.08151838922551\\
421	2.07590416247831\\
431	2.07145262760217\\
441	2.06556593406337\\
451	2.06134940682502\\
461	2.05594898976103\\
471	2.05068484068288\\
481	2.04720947861106\\
491	2.04259698679742\\
501	2.0387699947389\\
511	2.03462301212307\\
521	2.03116158050365\\
531	2.02724078122879\\
541	2.02413757986341\\
551	2.02112277846582\\
561	2.01759388204064\\
571	2.0138344736681\\
581	2.00979745928669\\
591	2.00625994588163\\
601	2.00324500670678\\
611	1.99967679682167\\
621	1.99670504249036\\
631	1.99429991376128\\
641	1.99188951524988\\
651	1.98909021759837\\
661	1.98631735063963\\
671	1.98362611874834\\
681	1.98100198384431\\
691	1.97876389908273\\
701	1.97564085626381\\
711	1.97309655839946\\
721	1.97033703682647\\
731	1.96823925429954\\
741	1.96579332747675\\
751	1.96414464534136\\
761	1.96234093381076\\
771	1.96093282597551\\
781	1.95873000973528\\
791	1.95684670158002\\
801	1.95422386170048\\
811	1.95209315755877\\
821	1.95052048114787\\
831	1.94895142726973\\
841	1.94642112049426\\
851	1.94448063266356\\
861	1.94266770865038\\
871	1.94104927644936\\
881	1.93980634257911\\
891	1.93846371593787\\
901	1.93685418071342\\
911	1.93528947997907\\
921	1.93413312141927\\
931	1.93270752486341\\
941	1.93145973302519\\
951	1.92924459124203\\
961	1.92760550378367\\
971	1.92621937885727\\
981	1.92467143733414\\
991	1.92326665954306\\
};
\addlegendentry{SPA-GPTD}

\addplot [color=mycolor2, line width=2.0pt]
  table[row sep=crcr]{%
51	2.86017726219722\\
61	2.77837535006655\\
71	2.72245418214734\\
81	2.68799149043882\\
91	2.65611013310295\\
101	2.62514979776032\\
111	2.58222009709459\\
121	2.5495944857644\\
131	2.519426833999\\
141	2.49224066689059\\
151	2.46959637206518\\
161	2.44238680541014\\
171	2.41656043258073\\
181	2.39824883351556\\
191	2.37874767114493\\
201	2.36213803349546\\
211	2.34292124601675\\
221	2.32347350413487\\
231	2.30709836527136\\
241	2.29354132347766\\
251	2.28200160474474\\
261	2.26896311036681\\
271	2.25627388595776\\
281	2.24196586862754\\
291	2.23263983138396\\
301	2.22161739113779\\
311	2.21092867722597\\
321	2.2007950035787\\
331	2.19211559769604\\
341	2.1833889726055\\
351	2.17430363870114\\
361	2.16641228977741\\
371	2.15941806980068\\
381	2.15273515245729\\
391	2.14454013764006\\
401	2.13751344190806\\
411	2.13100153791527\\
421	2.12434231897579\\
431	2.11886905216224\\
441	2.11203954320167\\
451	2.10687932933029\\
461	2.10060603872021\\
471	2.09450537376341\\
481	2.09019061543535\\
491	2.08479675583313\\
501	2.08020370925623\\
511	2.07532982829747\\
521	2.07115804440356\\
531	2.06656035902126\\
541	2.06278878924353\\
551	2.0591267539426\\
561	2.05498411489877\\
571	2.05063826001905\\
581	2.04604121930612\\
591	2.04195025229909\\
601	2.03839692871764\\
611	2.03431446238404\\
621	2.03083565338702\\
631	2.02792994672964\\
641	2.02503484605017\\
651	2.0217700725203\\
661	2.01854775271143\\
671	2.01541934839701\\
681	2.01237009202544\\
691	2.00971259566109\\
701	2.00619637116114\\
711	2.00326197195857\\
721	2.00012504613031\\
731	1.99765014127263\\
741	1.99484391192049\\
751	1.99283166280543\\
761	1.99067610082246\\
771	1.98892044945395\\
781	1.98639015483191\\
791	1.98418345973363\\
801	1.98125696060717\\
811	1.97882204810056\\
821	1.97694551452727\\
831	1.97507887630434\\
841	1.97227173226648\\
851	1.97005371657229\\
861	1.9679679804086\\
871	1.96608027821709\\
881	1.96456862080574\\
891	1.96296524914895\\
901	1.96110483938332\\
911	1.95929326459275\\
921	1.95789070602308\\
931	1.95622746474013\\
941	1.95474545371789\\
951	1.95231215827403\\
961	1.95045290359946\\
971	1.94884731862891\\
981	1.9470871758854\\
991	1.94547247692454\\
};
\addlegendentry{OS-GPTD}

\addplot [color=mycolor3,line width=2.0pt]
  table[row sep=crcr]{%
51	2.73147391888095\\
61	2.64304032983278\\
71	2.58505858110064\\
81	2.54220391756816\\
91	2.50444224361558\\
101	2.47042887203192\\
111	2.43045998504878\\
121	2.40102420949761\\
131	2.37177919665115\\
141	2.34759193171667\\
151	2.3268680035572\\
161	2.30392084678747\\
171	2.28222921557063\\
181	2.26643031427935\\
191	2.24969448015958\\
201	2.23526291208346\\
211	2.21956570033721\\
221	2.20284266498477\\
231	2.18914463422001\\
241	2.1771767792614\\
251	2.16738569940771\\
261	2.15670358327834\\
271	2.14646322356492\\
281	2.13492395755157\\
291	2.12789234195477\\
301	2.11895797160334\\
311	2.11057302017464\\
321	2.10212491080126\\
331	2.09547094881665\\
341	2.08860025275181\\
351	2.08150206981116\\
361	2.07468535680798\\
371	2.069346358527\\
381	2.06437393794685\\
391	2.05736715446545\\
401	2.05142941627217\\
411	2.04626530494886\\
421	2.0406384210919\\
431	2.03650375088945\\
441	2.03081285026117\\
451	2.02678228980506\\
461	2.02179292341491\\
471	2.01686217588626\\
481	2.01342277257822\\
491	2.00910397647063\\
501	2.00542931647549\\
511	2.00148794764511\\
521	1.99816820561624\\
531	1.99442226282818\\
541	1.99163166653447\\
551	1.98877182737654\\
561	1.9855536632698\\
571	1.98206548430055\\
581	1.97845171027976\\
591	1.97517993002911\\
601	1.97235993126304\\
611	1.96905428565917\\
621	1.96622986650623\\
631	1.96403211118998\\
641	1.96176292858648\\
651	1.95908150812347\\
661	1.95666870662742\\
671	1.95415898898847\\
681	1.95173601503827\\
691	1.94975356029962\\
701	1.94699589561485\\
711	1.94473677800178\\
721	1.94225812548997\\
731	1.94029107659482\\
741	1.93801927867812\\
751	1.93656799462174\\
761	1.93497229380293\\
771	1.93362660074718\\
781	1.93169840041263\\
791	1.92991025393856\\
801	1.9274627709755\\
811	1.92549548121453\\
821	1.92413896681538\\
831	1.9227703651641\\
841	1.92045155589202\\
851	1.91859156618211\\
861	1.91699608824281\\
871	1.91549533341034\\
881	1.9143858342211\\
891	1.91315438725519\\
901	1.91171327563419\\
911	1.91030985652697\\
921	1.90930504007179\\
931	1.90802810310744\\
941	1.90691020988898\\
951	1.90487007540768\\
961	1.9034146933602\\
971	1.90219284768006\\
981	1.90083852600369\\
991	1.8995459671033\\
};
\addlegendentry{OS-EGPTD}

\end{axis}
\end{tikzpicture}%
}
	\caption{Average Bellman error on {\bf Random Walk}.}
	\label{fig:RW_Be}
\end{figure}

\section{Experiments}
In this section, our novel OS-(E)GPTD schemes are tested using two simulated policy evaluation benchmarks. To better assess their performance, we compare them with the online sparsification approach based GPTD devised in
\cite{engel2003bayes}, henceforth abbreviated as SPA-GPTD. Performance metrics include the average cumulative long-term reward prediction error $\sqrt{\sum_{\tau=1}^t e_\tau^2/t }$ and Bellman error $\sqrt{\sum_{\tau=1}^t l_\tau(\hat{\bbtheta}_{\tau-1})/t}$ up to slot $t$, as well as the average runtime. The kernel dictionary of OS-EGPTD comprises three standardized Gaussian kernels, whose characteristic lengthscales ($\sigma$'s) are $\mathcal{K}:= \{0.1, 1, 10 \}$, while both OS-GPTD and SPA-GPTD employ a single Gaussian kernel with $\sigma \in \mathcal{K}$ specified below. The reported results are averaged over $100$ trajectories.

\begin{figure}[t]
	\centering{\input{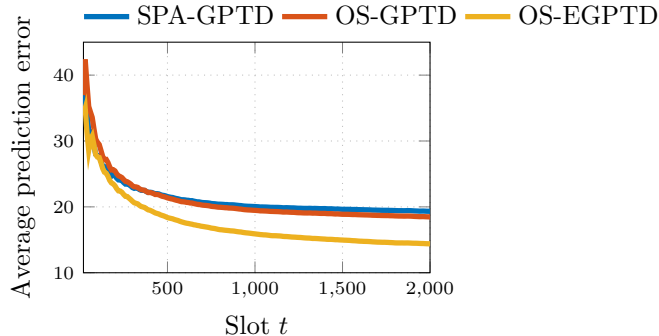}}
	\caption{Average prediction error on {\bf Puddle World}.}
	\label{fig:PW_Ve}
\end{figure}

\begin{figure}[t]
	\centering{\input{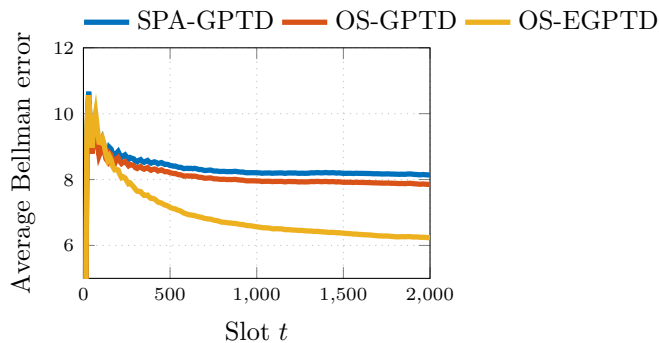}}
	\caption{Average Bellman error on {\bf Puddle World} .}
	\label{fig:PW_Be}
\end{figure}

\noindent {\bf Random Walk.} This problem is a variant of the Hall problem introduced in \cite{baird1995residual}, and also used in \cite{li2008worst,sun2015online}. The state space consists of $N=50$ states, which are linked as a ring without terminal states. Per slot $t$, the current state transitions clockwise to the next state with probability $1$.  Each state is associated with a vector $\mathbf{s}_i\in\mathbb{R}^{10}$, $i=1,\ldots,50$, randomly generated from the standard normal distribution. Instantaneous rewards $r_t$ are uniformly sampled from $[-3, 3]$, and each trajectory contains at most $T=1,000$ slots. For OS-GPTD and SPA-GPTD, the single GP prior uses a standardized Gaussian kernel with $\sigma = 0.1$. The discount factor $\gamma$ is set to $0.75$, and $\sigma_n^2=0.01$.

\begin{figure}[t]
	\centering{\definecolor{mycolor1}{rgb}{0.00000,0.44700,0.74100}%
\definecolor{mycolor2}{rgb}{0.85000,0.32500,0.09800}%
\definecolor{mycolor3}{rgb}{0.92900,0.69400,0.12500}%

\begin{tikzpicture}
\begin{axis}
[%
width=1.\mywidth,
height=1\myheight,
at={(0\mywidth,0\myheight)},
scale only axis,
bar shift auto,
log origin=infty,
xmin=-1.2,
xmax=8.2,
xtick={1,6},
xticklabels={{\texttt{RandomWalk},\texttt{PuddleWorld}}
},
ymode=log,
ymin=0,
ymax=80,
yminorticks=true,
ylabel style={font=\color{white!15!black}},
ylabel={Normalized Runtime},
ylabel near ticks,
axis background/.style={fill=white},
ymajorgrids,
yminorgrids,
legend columns=4,
legend style={
 at={(0,1.015)},
 anchor=south west, legend cell align=left, align=left, draw=none
 , font=\legendfontsize}
]

\addplot[ybar, bar width=1, fill=mycolor1, draw=mycolor1, area legend] table[row sep=crcr]
{%
1  5.3842\\
6 80.4856\\
};
\addlegendentry{SPA-GPTD}

\addplot[ybar, bar width=1, fill=mycolor3, draw=mycolor3, area legend] table[row sep=crcr]
{%
 1  1\\
    6  1\\
};
\addlegendentry{OS-EGPTD}

\addplot[ybar, bar width=1, fill=mycolor2, draw=mycolor2, area legend] table[row sep=crcr] 
{%
 1 0.1014\\
 6 0.1119\\
};
\addlegendentry{OS-GPTD}

\end{axis}
\end{tikzpicture}%
}
	\caption{Average CPU times.}
	\label{fig:RW_RT}
\end{figure}
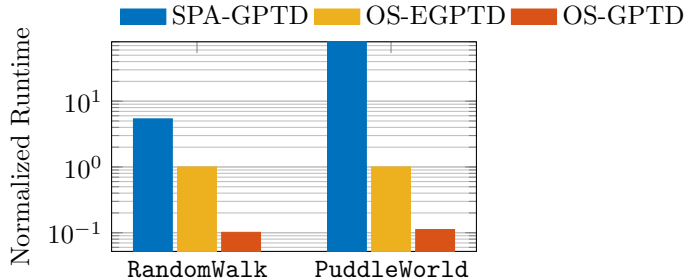

\noindent {\bf PuddleWorld} 
was adapted from \cite{sutton2018reinforcement} with its state space being a unit square with `puddles.' The agent’s state is represented by its $x$ and $y$ coordinates. In each episode, the starting state is uniformly sampled in the region $[0,0.2]\times [0,0.2]$. The policy per step selects `go north or east' with probability $0.5$. The terminal region is a circle with radius $0.1$ centered at $(1,1)$. The reward $r_t$ is $-1$ if the agent does not step in a puddle, and decreases to $-2$ as the agent steps into the puddles or boundaries. 
Both OS-GPTD and SPA-GPTD employ a Gaussian kernel with $\sigma = 1$, while $\sigma_n^2 = 0.001$ and $\gamma=0.7$. 

As evident from Figs.~\ref{fig:RW_Ve}-\ref{fig:PW_Be}, our OS-EGPTD enjoys lower prediction and Bellman errors than single GP-based alternatives. Although achieving similar estimation performances, the RF-based OS-GPTD scheme is much more computationally efficient than SPA-GPTD as shown in Fig.~\ref{fig:RW_RT}. Additional tests to validate the regret bound in Theorem \ref{th:1} can be found in the supplementary file.

\section{Conclusions}
Building on the GPTD framework for value estimation in RL, this paper developed OS-(E)GPTD schemes,
leveraging RF-based kernel approximants using an (ensemble of) GP prior(s) to effect online function estimation along with performance quantification. To account for violations of the GPTD modeling assumptions in practice, worst-case analyses have been performed in terms of both the Bellman error as well as the long-term reward prediction error, that benchmark the performance of online function estimators produced by OS-GPTD relative to the best fixed one with batch data in hindsight. Numerical tests corroborated the merits of the novel 
OS-(E)GPTD
schemes.

\bibliographystyle{IEEEtranS}
\bibliography{learning}



\appendix

\section{Proof of Lemma \ref{le:r1}}
\label{pf:le}

It will be useful to rewrite the update in \eqref{eq:SGP_up} as
\begin{equation}
\hat{\bbtheta}_{t} = \hat{\bbtheta}_{t-1}  - \frac{1}{2\sigma_n^2} \mathbf{J}_{t}^{-1} \nabla_{t|t-1}  \label{eq:update_new}
\end{equation}
where the gradient of the Bellman error at slot $t$ evaluated at $\hat{\bbtheta}_{t-1}$ is 
\begin{equation}
 \nabla_{t|t-1} := 2\mathbf{h}_{t} (\mathbf{h}_{t}^\top \hat{\bbtheta}_{t-1} - r_{t})\label{eq:BE_GD}  
\end{equation}
and $\mathbf{J}_t := \bbSig_t^{-1}$ denotes the information matrix, which starting from 
$\mathbf{J}_0 : = \sigma_{\theta}^{-2}\mathbf{I}_{2D}$, can be propagated using
\begin{equation}
\mathbf{J}_t = \mathbf{J}_{t-1} + \sigma_n^{-2} \mathbf{h}_t \mathbf{h}_t^{\top}. \label{eq:J_iter}
\end{equation}	
Invoking Taylor's expansion, we can write 
\begin{align}
&l_t(\hat{\bbtheta}_{t-1})-l_t(\bbtheta^{*}) \nonumber\\
=&  -\!\nabla_{t|t-1}^\top (\bbtheta^{*}\!- \hat{\bbtheta}_{t-1}\!) \!-\! (\bbtheta^{*}\!- \hat{\bbtheta}_{t-1}\!)^\top   \mathbf{h}_t \mathbf{h}_t^\top \!(\bbtheta^{*} \!- \hat{\bbtheta}_{t-1}\!) \nonumber\\
 \overset{(a)}{=} & -\!\nabla_{t|t-1}^\top (\bbtheta^{*}\! -\! \hat{\bbtheta}_{t-1}) \!-\! \sigma_n^2(\bbtheta^{*}\! -\! \hat{\bbtheta}_{t-1}\!)^\top \mathbf{J}_t(\bbtheta^{*} \!-\! \hat{\bbtheta}_{t-1}) \nonumber\\
&   + \sigma_n^2(\bbtheta^{*}- \hat{\bbtheta}_{t-1}\!)^\top \mathbf{J}_{t-1}(\bbtheta^{*} - \hat{\bbtheta}_{t-1}) \nonumber
\end{align}
where $(a)$ utilizes  $\mathbf{h}_t \mathbf{h}_t^\top = \sigma_n^2 (\mathbf{J}_t - \mathbf{J}_{t-1})$ from \eqref{eq:J_iter}. The cumulative regret $\mathcal{R}_1(T)$ now becomes
\begin{align}
\mathcal{R}_1&(T)= \sum_{t =1}^T l_t(\hat{\bbtheta}_{t-1})- \sum_{t =1}^Tl_t(\bbtheta^{*})  \nonumber\\
 =  &\sum_{t=1}^{T} \Big[\!-\!  \nabla_{t|t-1}^\top (\bbtheta^{*} \!- \hat{\bbtheta}_{t-1}) \! +\!\sigma_n^2 (\bbtheta^{*}\! -\! \hat{\bbtheta}_{t})^\top \!\mathbf{J}_{t } (\bbtheta^{*}\! -\! \hat{\bbtheta}_{t})  \nonumber\\
& \quad -\sigma_n^2 (\bbtheta^{*} - \hat{\bbtheta}_{t-1})^\top \mathbf{J}_{t } (\bbtheta^{*} 
-\hat{\bbtheta}_{t-1})  \Big] + \bbtheta^{*\top} \mathbf{J}_{0 } \bbtheta^{*} \nonumber\\
&  \quad  - (\bbtheta^{*} - \hat{\bbtheta}_{T})^\top \mathbf{J}_{T } (\bbtheta^{*} - \hat{\bbtheta}_{T})  . \label{eq:R1_1}
\end{align}
On the other hand, the update in \eqref{eq:update_new} confirms that
\begin{align}
&(\bbtheta^{*}-\hat{\bbtheta}_{t})^\top \mathbf{J}_t (\bbtheta^{*}-\hat{\bbtheta}_{t})   \nonumber \\
& =   \!\bigg(\!\bbtheta^{*}\!-\!\hat{\bbtheta}_{t-1} \!+\! \frac{\mathbf{J}_{t}^{-1} \nabla_{t|t-1} }{2\sigma_n^2}\!\! \bigg)^{\!\top}\!\! \!\mathbf{J}_t\!\bigg(\!\bbtheta^{*}\!-\!\hat{\bbtheta}_{t-1}\!+\! \frac{\mathbf{J}_{t}^{-1} \nabla_{t|t-1}}{2\sigma_n^2}  \!\bigg)                \nonumber \\
&= (\bbtheta^{*}\!-\!\hat{\bbtheta}_{t-1})^\top \mathbf{J}_t (\bbtheta^{*}\!-\!\hat{\bbtheta}_{t-1}) +\frac{1}{\sigma_n^2} \nabla_{t|t-1}^\top  (\bbtheta^* - \hat{\bbtheta}_{t-1}) \nonumber\\
&\quad  +\frac{1}{4\sigma_n^4} \nabla_{t|t-1}^\top  \mathbf{J}_t^{-1} \nabla_{t|t-1} \nonumber
\end{align}
substituting which into \eqref{eq:R1_1}, yields
\begin{align}
\mathcal{R}_1 (T) &= \sum_{t=1}^T \frac{1}{4\sigma_n^2}  \nabla_{t|t-1}^\top  \mathbf{J}_t^{-1} \nabla_{t|t-1} + \bbtheta^{*\top} \mathbf{J}_{0 }\bbtheta^{*} \nonumber\\   
&  \quad- (\bbtheta^{*} - \hat{\bbtheta}_{T})^\top \mathbf{J}_{T } (\bbtheta^{*} - \hat{\bbtheta}_{T}) 	\nonumber \\
& \overset{(a)}{\leq}    \sum_{t=1}^T \frac{1}{4\sigma_n^2}  \nabla_{t|t-1}^\top  \mathbf{J}_t^{-1} \nabla_{t|t-1}  + \frac{\| \bbtheta^{*}\|_2^2 }{2 \sigma_{\theta}^2} \nonumber\\
&\overset{(b)}{\leq}    \sum_{t=1}^T \frac{4B_e^2}{4\sigma_n^2}  \mathbf{h}_t^\top  \mathbf{J}_t^{-1} \mathbf{h}_t  + \frac{\| \bbtheta^{*}\|_2^2 }{2 \sigma_{\theta}^2} \nonumber\\
& \overset{(c)}{\leq}  2D B_e^2 \log \!\left(\frac{(1+\gamma)^2\sigma_{\theta}^2}{\sigma_n^2} T +1 \right) + \frac{\| \bbtheta^{*}\|_2^2 }{2 \sigma_{\theta}^2} \nonumber
\end{align}
where $(a)$ follows from $(\bbtheta^{*} - \bbtheta_{T})^\top \mathbf{J}_{T } (\bbtheta^{*} - \bbtheta_{T}) \geq 0$ and $\mathbf{J}_0=\sigma_{\theta}^{-2}\mathbf{I}_{2D}$, $(b)$ leverages the gradient evaluation in \eqref{eq:BE_GD}, and $(c)$ relies on a standard result in \cite[Appendix 2]{hazan2007logarithmic} with $\|\mathbf{h}_{t}\| \leq 1+\gamma$ and $|r_t - \mathbf{h}_t^\top \hat{\bbtheta}_{t-1}| \leq B_e$ under our assumptions. This concludes the proof of Lemma \ref{le:r1}.

\section{Proof of Theorem 1}
For a given shift-invariant, standardized kernel $\bar{\kappa}^s$, the maximum point-wise error of the RF kernel approximant can be uniformly bounded with probability at least 
$1- \frac{2^8\sigma_{\zeta}^2}{\epsilon^2} \,{ e}^{-\frac{D\epsilon^2}{4d+8} }$	
	by \cite{rahimi2008random}
	\begin{equation}
	\sup_{\mathbf{s}_i, \mathbf{s}_j \in \mathcal{S}}\; \left|\bbphi^{\top}_{\zeta}(\mathbf{s}_i)\bbphi_{\zeta}(\mathbf{s}_j) - \bar{\kappa} (\mathbf{s}_i, \mathbf{s}_j)\right| < \epsilon \label{eq:kernel_approx}
	\end{equation}
	where $\epsilon$ is a given constant, $D$ is the number of RF vectors, $d$ is the dimension of $\mathbf{s}$, and $\sigma_{\zeta}^2 = \mathbb{E}_{\pi_{\bar{\kappa}}}[\| \bbzeta\|^2]$ is the second-order moment of the RF vector $\bbzeta$.
	
	The optimal function estimator in $\mathcal{H}$ induced by $\kappa$ is given by
	$v^{*} (\mathbf{s}):= \sum_{t = 1}^T \hat \alpha_{t} \kappa (\mathbf{s}, \mathbf{s}_t) =  \sigma_{\theta}^2\sum_{t = 1}^T \hat \alpha_{t} \bar{\kappa} (\mathbf{s}, \mathbf{s}_t) $; and its RF-based approximant is $\check{v}^* (\mathbf{s}):= \bbphi^{\top}_{\zeta}(\mathbf{s})\bbtheta^*$ with $\bbtheta^*:=\sigma_{\theta}^2\sum_{t = 1}^{T} \hat \alpha_{t} \bbphi_{\zeta} (\mathbf{s}_t)$. We then have that	\begin{align}
	&\big| l_t(\check{v}^*) - l_t ( v^* )   \big|  \nonumber \\
\overset{(a)}{\leq} 	& \, 2B_e \big|\check{v}^*(\mathbf{s}_{t-1})-\gamma\check{v}^*(\mathbf{s}_{t}) - [v^*(\mathbf{s}_{t-1})-\gamma v^*(\mathbf{s}_{t}) ] \big| \nonumber\\
	\overset{}{\leq} & 2B_e\big[\big|\check{v}^*(\mathbf{s}_{t-1})- v^*(\mathbf{s}_{t-1})\big|+ \gamma\big|\check{v}^*(\mathbf{s}_{t}) - v^*(\mathbf{s}_{t}) \big|\big] \nonumber\\
	 \overset{}{=}  &\, 2B_e\!\sigma_{\theta}^2 \Big[\Big|\sum_{t'=1}^{T} \hat\alpha_{t'}\big ( \bbphi_{\zeta}^{\top}\! (\mathbf{s}_{t-1}) \bbphi_{\zeta}(\mathbf{s}_{t'}) -  \bar{\kappa} (\mathbf{s}_{t-1}, \mathbf{s}_{t'})\big )  \Big| \nonumber \\
	&  +  \gamma  \Big|\sum_{t'=1}^{T} \hat\alpha_{t'}\big ( \bbphi_{\zeta}^{\top} (\mathbf{s}_{t}) \bbphi_{\zeta}(\mathbf{s}_{t'}) -  \bar{\kappa} (\mathbf{s}_{t}, \mathbf{s}_{t'}) \big )  \Big|\Big] \nonumber \\
	\overset{
	}{\leq} 
	& 2B_e\sigma_{\theta}^2 \sum_{t'=1}^{T} |\hat\alpha_{t'}| \Big(\big|\bbphi_{\zeta}^{\top} (\mathbf{s}_{t-1}) \bbphi_{\zeta}(\mathbf{s}_{t'}) -   \bar{\kappa} (\mathbf{s}_{t-1}, \mathbf{s}_{t'})   \big|  \nonumber\\
	&      + \gamma \big|\bbphi_{\zeta}^{\top} (\mathbf{s}_t) \bbphi_{\zeta}(\mathbf{s}_{t'}) -   \bar{\kappa} (\mathbf{s}_t, \mathbf{s}_{t'})   \big| \Big)
	\end{align}
where $(a)$ follows because $l_t(v)$ is convex wrt $v({\mathbf{s}_t}) - \gamma v({\mathbf{s}_{t+1}})$, with gradient $2(v({\mathbf{s}_t})-\gamma v({\mathbf{s}_{t+1}})+r_t )$, bounded by $2B_e$. Combining with \eqref{eq:kernel_approx}, we deduce that
	\begin{equation}
	 | l_t (\check{v}^*) -  l_t ( v^* )   |  \leq
	2(1+\gamma)B_e C\epsilon,~~~ \ {\rm w.h.p.}  \nonumber
	\end{equation}
	where $C:=\sum_{t = 1}^T \sigma_{\theta}^2 | \hat\alpha_{t}|$. It thus holds that
	\begin{equation}
\sum_{t = 1}^T l_t (\check{v}^*) \!-\! \sum_{t = 1}^Tl_t ( v^* ) \leq 2(1+\gamma)\epsilon B_e C T ,\,  {\rm w.h.p.} \label{eq:lambda_RF_RKHS_2}
	\end{equation}
On the other hand, the uniform convergence bound in \eqref{eq:kernel_approx} along with as2) imply that
	\begin{equation}
	\underset{\mathbf{s}_t, \mathbf{s}_{t'} \in \mathcal{S}}{\sup }~~\bbphi^{\top}_{\zeta} (\mathbf{s}_t) \bbphi_{\zeta}(\mathbf{s}_{t'}) \leq 1+ \epsilon, \quad {\rm w.h.p.} \label{eq:sup_phi2}
	\end{equation}
	which leads to 
	\begin{align}
	\|\bbtheta^{*} \|^2&\!:=  \Big\|\sigma_{\theta}^2\sum_{t = 1}^T \hat\alpha_{t} \bbphi_{\zeta} (\mathbf{s}_t)  \Big\|^2 \label{eq:theta_bound}  \\
	& =  \sigma_{\theta}^4\sum_{t = 1}^T\sum_{t' = 1}^T \hat\alpha_{t} \hat\alpha_{t'}\bbphi^{\top}_{\zeta} (\mathbf{s}_t)\bbphi_{\zeta} (\mathbf{s}_{t'}) \nonumber\\
	&\leq (1+\epsilon)C^2    .  \nonumber
	\end{align}
Putting together the bounds in 
\eqref{eq:theta_bound} and \eqref{eq:lambda_RF_RKHS_2}, along with Lemma \ref{le:r1}, we arrive at \eqref{eq:theorem1}, which completes the proof of Theorem \ref{th:1}.

\section{Proof of Theorem 2}
Recall first that \cite[Thm. 3.3]{sun2015online} 
\begin{align}\vspace{-.2cm}
& (1-\gamma)^2\sum_{t=1}^T e_t^2 \leq 
2(1+\gamma)^2 \sum_{t=1}^T e_t^{*2}+ C'  \label{eq:T2_proof}\\
& 
  +\! 2\gamma^2 \sum_{t=1}^T\! (\hat{\check v}_{t}\big(\mathbf{s}_t)- \hat{\check v}_{t-1}(\mathbf{s}_{t})\big)^2 \!+\! 2\sum_{t=1}^T\big (l_t(\hat{\bbtheta}_{t-1})\!-\! l_t (v^{*})\big)\nonumber
\end{align}
where $C'$ is a constant. Using Theorem \ref{th:1}, it follows that
\begin{equation}
\lim_{T\rightarrow \infty} \frac{1}{T}\sum_{t=1}^T\big (l_t(\hat{\bbtheta}_{t-1})- l_t (v^{*})\big)=0. \label{eq:BE_noregret}
\end{equation}	
To deduce \eqref{eq:PE_bound} from \eqref{eq:T2_proof}, it suffices to show that
\begin{equation}
\lim_{T\rightarrow \infty} \frac{1}{T}\sum_{t=1}^T \big(\hat{\check v}_{t}(\mathbf{s}_t)\!-\! \hat{\check v}_{t-1}(\mathbf{s}_{t})\big)^2=0 \label{eq:OS}.
\end{equation}
which is designated as the \emph{online stability} condition in e.g., \cite{sun2015online}. 
To establish \eqref{eq:OS}, we need the following inequality
\begin{align}
&(\hat{\check v}_{t}(\mathbf{s}_t)- \hat{\check v}_{t-1}(\mathbf{s}_{t}))^2  = (\bbphi_{\zeta}^\top (\mathbf{s}_t) (\hat{\bbtheta}_t - \hat{\bbtheta}_{t-1}))^2 \nonumber\\
&\overset{(a)}{\leq} \! \|\bbphi_{\zeta}(\mathbf{s}_t)\|^2 \|\hat{\bbtheta}_{t}- \hat{\bbtheta}_{t-1}\|^2  \overset{}{=}\! \left\| \frac{\bbSig_{t-1} \mathbf{h}_{t}(r_{t} \!-\! \mathbf{h}_{t}^\top \hat{\bbtheta}_{t-1})}{\mathbf{h}_{t}^\top\bbSig_{t}\mathbf{h}_{t}\! +\! \sigma_n^2 } \right\|^2 \nonumber \\
& \overset{(b)}{\le } \frac{B_e^2}{\sigma_n^2} \frac{\mathbf{h}_{t}^\top \bbSig_{t-1}\bbSig_{t-1} \mathbf{h}_{t}  }{\mathbf{h}_{t}^\top\bbSig_{t}\mathbf{h}_{t}\! +\! \sigma_n^2}    =\frac{B_e^2}{\sigma_n^2}  {\rm Tr}\! \left(\frac{ \bbSig_{t-1} \mathbf{h}_{t} \mathbf{h}_{t}^\top \bbSig_{t-1}}{\mathbf{h}_{t}^\top\bbSig_{t-1}\mathbf{h}_{t}\! +\! \sigma_n^2}  \right) \nonumber \\
& \overset{(c)}{=} \frac{B_e^2}{\sigma_n^2} {\rm Tr} (\bbSig_{t-1} - \bbSig_{t}) \label{eq:stability}
\end{align}
where $(a)$ arises from the Cauchy-Schwarz inequality; $(b)$ is due to as1) and $\mathbf{h}_{t}^\top\bbSig_{t}\mathbf{h}_{t} >0$; and the identity $(c)$ employs the update of the covariance matrix in \eqref{eq:cov_up}.

Summing \eqref{eq:stability} from $t=1$ to $T$, yields 	 \vspace{-.2cm}
\begin{equation}
\vspace{-.2cm}
\sum_{t=1}^T\big (\hat{\check v}_{t}(\mathbf{s}_t)- \hat{\check v}_{t-1}(\mathbf{s}_{t})\big)^2 \leq \frac{B_e^2}{\sigma_n^2}   {\rm Tr} (\bbSig_{0} - \bbSig_{T})   
\end{equation}
which, in conjunction with \eqref{eq:BE_noregret}, validates \eqref{eq:PE_bound}, thus completing the proof of Theorem \ref{th:2}.

\section{Additional tests}
\label{addtest}

\begin{figure}[h]
	\centering{\input{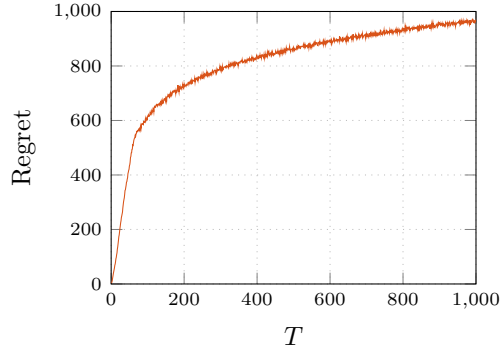}}
	\caption{Regret of OS-GPTD vs. $T$ on the {\bf Random Walk} problem.}
	\label{fig:RW_OSGP_Regret}
\end{figure}

\end{document}